\documentclass{article}
\usepackage{graphicx}
\usepackage{indentfirst,csquotes}

\usepackage{amssymb,amsthm,amsmath}
\usepackage{xcolor,paralist,hyperref,titlesec,fancyhdr,etoolbox}

\hypersetup{ colorlinks=true, linkcolor=black, filecolor=black, urlcolor=black }

\usepackage{lipsum}

\usepackage[utf8]{inputenc} 
\usepackage[T1]{fontenc}    
\usepackage{hyperref}       
\usepackage{url}            
\usepackage{booktabs}       
\usepackage{amsfonts}       
\usepackage{nicefrac}       
\usepackage{microtype}      
\usepackage{lipsum}
\usepackage{fancyhdr}       
\usepackage{graphicx}       
\graphicspath{{media/}}     

\usepackage{svg}
\usepackage{lscape}
\usepackage{supertabular}
\usepackage{booktabs}
\usepackage{makecell}

\usepackage[numbers]{natbib}
\usepackage{longtable}
\usepackage{caption}
\usepackage{pdflscape}
\usepackage{authblk}
\hypersetup{
    colorlinks=true,
    linkcolor=blue,
    citecolor=blue,
    urlcolor=cyan
    }
\graphicspath{{Fig/}}

\begin{document}
\title{Language modelling techniques for analysing the impact of human genetic variation}
\author[1$\ast$]{Megha Hegde}
\author[1]{Jean-Christophe Nebel}
\author[1$\ast$]{Farzana Rahman}
\affil{School of Computer Science and Mathematics, Kingston University London}

\maketitle

\begin{abstract}
Interpreting the effects of variants within the human genome and proteome is essential for analysing disease risk, predicting medication response, and developing personalised health interventions. Due to the intrinsic similarities between the structure of natural languages and genetic sequences, natural language processing techniques have demonstrated great applicability in computational variant effect prediction. In particular, the advent of the Transformer has led to significant advancements in the field. However, Transformer-based models are not without their limitations, and a number of extensions and alternatives have been developed to improve results and enhance computational efficiency. This review explores the use of language models for computational variant effect prediction over the past decade, analysing the main architectures, and identifying key trends and future directions.
\end{abstract} 

\bigskip

\section{Introduction}
Understanding the impact of genetic variants is crucial for unravelling gene regulation mechanisms and disease causality. As we enter the era of personalised medicine, it has become of great interest to understand how an individual's genetic makeup can impact their risk of developing a particular disease or their response to a specific treatment or medication \citep{karki2015defining, goetz2018personalised}.

Any change in a coding region can directly affect the function of the associated protein, hence certain gene mutations can be linked with specific diseases. While Mendelian (monogenic) diseases, such as cystic fibrosis and haemophilia, are caused by mutations in a single gene \citep{jamuar2015clinical, rahit2020genetic}, polygenic diseases, including many cancers \citep{castro2015mini, jia2020evaluating}, result from combinations of mutations \citep{lvovs2012polygenic, visscher2021discovery}. Variation in the non-coding region of the genome is more challenging to interpret than that in the coding region, as variants impact disease-related genes by altering processes such as transcription, chromatin folding, or histone modification \citep{zhang2015noncoding, moyon2022classification}.

The advent of next-generation sequencing technology has made ever more genomic data available for analysis \citep{pereira2020bioinformatics}. Large-scale studies have shown a high degree of genetic variation between humans, with the 1000 Genome Project Consortium postulating the existence of at least 88 million unique variants across the global population \citep{10002015global}. Therefore, computational approaches, particularly those utilising machine learning, have come into favour due to their capability to process and analyse large datasets \citep{liu2022computational, bromberg2024variant}. In particular, language models have demonstrated notable efficacy on such tasks, due to their ability to capture the dependencies between different parts of genetic sequences and take into account contextual information \citep{ji2021dnabert}.
Indeed, linguistic metaphors, from alphabets to grammars, have been readily used to describe the molecular world since the discovery of the structure of DNA in the 1950s \citep{brendel1984nucleic, brendel1986linguistics}. For instance, as genetic sequences are comprised of nucleotides or amino acids represented as letters, the sequences themselves can be represented as strings of letters, and processed in a way that is analogous to human language \citep{searls_1992_the_linguistics_of_dna, solan2005unsupervised}.

Although Noam Chomsky formed the basis of modern language modelling in the 1950s \citep{chomsky1956three}, the field has advanced considerably over the decades. A pivotal point was the development of the Transformer in 2017 \citep{vaswani2017attentionisallyouneed}; which sparked a discernible shift towards the use of so-called large language models (LLMs) to solve a plethora of language modelling tasks in bioinformatics, including variant effect prediction \citep{tang2023building, bromberg2024variant}. These LLMs are Transformer-based models with billions of parameters, trained on large corpora of sequence data, and have been favoured due to their ability to accurately model long-range dependencies within sequences \citep{zhao2023survey, shanahan2024talking}.

Large language models have been used extensively in bioinformatics, and many excellent reviews have been published on several aspects \citep{zhang2023applications, sarumi2024large, tian2024opportunities}, however, none have yet focused on large language models for variant effect prediction. Hence, this review addresses this gap by first presenting an introduction to variant effect prediction and biological language modelling, before entering an in-depth exploration of language models applied to the prediction of effects of genetic variations within DNA, RNA, and protein sequences. Following a brief presentation of the history of language modelling, in line with the rapid advancement in the field, the core of the review covers models produced since the inception of the Transformer in 2017. This review focuses on variants within the human genome, and their impacts on disease causality, however, models trained on multi-species data are also considered.

\section{Background}
As this manuscript details the applications of language modelling to variant effect prediction tasks, this section provides a brief introduction to both aspects - variant effect prediction and natural language processing - to set out the main problems in the field, and the technologies that can be used to address them.

\subsection{Variant Effect Prediction}
Uncovering the associations between genetic variants and human diseases necessitates an understanding of the many different possible types of variants. The variants most commonly explored in the field are single base-pair substitutions, referred to as single-nucleotide polymorphisms (SNPs). Still, a small number of models have been developed to analyse the combined effect of several SNPs \citep{castro2015mini, amadeus2021design}. While several single base-pair substitutions can co-occur independently, they can also occur as a single event; in such cases, they are referred to as multiple base-pair substitutions \citep{hampsey1988multiple}. However, there is no evidence they have been addressed in the literature.
In addition to substitutions, two other significant forms of variation are insertions and deletions, collectively known as indels. Insertion refers to the case where additional nucleotides are inserted into a genetic sequence, while deletion refers to the case where nucleotides are deleted from such a sequence. Similarly to substitutions, these events can occur across single or multiple nucleotides. While some papers have investigated indel effect prediction \citep{luo2024interpretable}, this has been explored to a substantially lesser extent than substitutions.

\begin{figure}
    \centering
    \includegraphics[width=\linewidth]{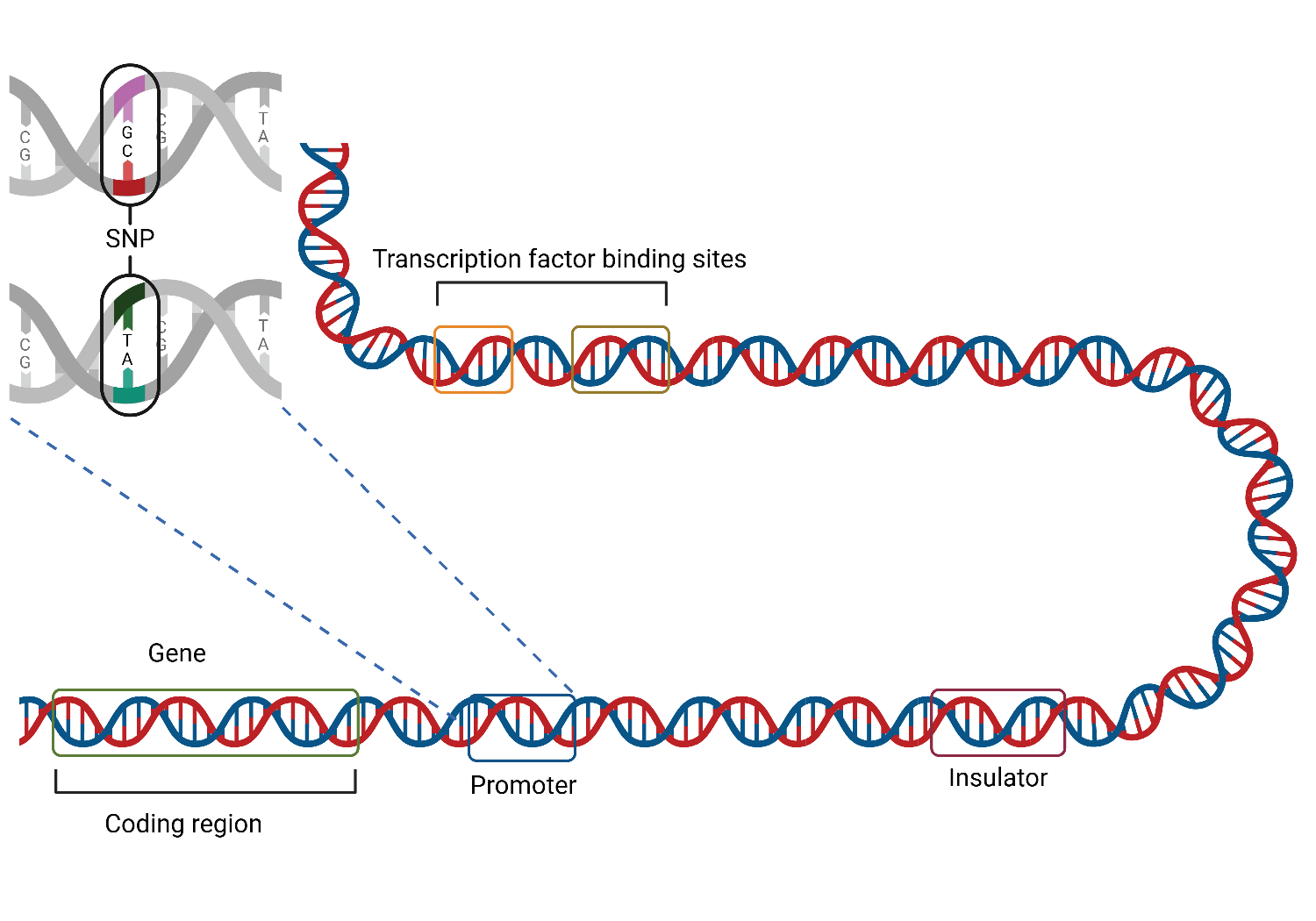}
    \caption{Illustration of coding vs non-coding DNA, and a SNP in a promoter region, for a eukaryotic cell. Non-coding DNA consists of transcription factors, such as promoters, and transcription factor binding sites. Promoters drive the initiation of transcription \citep{andersson2020determinants}. Other cis-regulatory elements (CREs) include enhancers and silencers, which positively and negatively regulate gene expression, respectively. Insulators are an additional type of CRE, which interact with nearby CREs and can block distal enhancers, or regulate chromatin interactions \citep{west2002insulators}. Created in BioRender. Hegde, M. (2024) https://BioRender.com/e16b233.\\
    \textbf{Alt text:} Illustration of DNA, with coding region (gene) and key non-coding regions (promoter, insulator, transcription factor binding sites) highlighted and labelled, and a visual representation of a SNP.}
    \label{fig:dna}
\end{figure}

Existing work focuses largely on variants within genes, which code for proteins. However, these protein-coding regions comprise less than 2\% of the human genome \citep{lander2011initial}. As illustrated in Figure \ref{fig:dna}, variants can also occur in the non-coding regions of the genome, including in regulatory elements such as promoters and enhancers. In fact, 90\% of disease-associated variants identified by genome-wide association studies have mapped to non-coding regions, and the majority of these remain unannotated \citep{gaulton2023interpreting}. Hence, the discovery of non-coding variant effects remains a largely untapped source of potential knowledge that could aid in illuminating human disease mechanisms.

\subsection{Natural Language Processing}

\begin{figure}
    \centering
    \includegraphics[width=\linewidth]{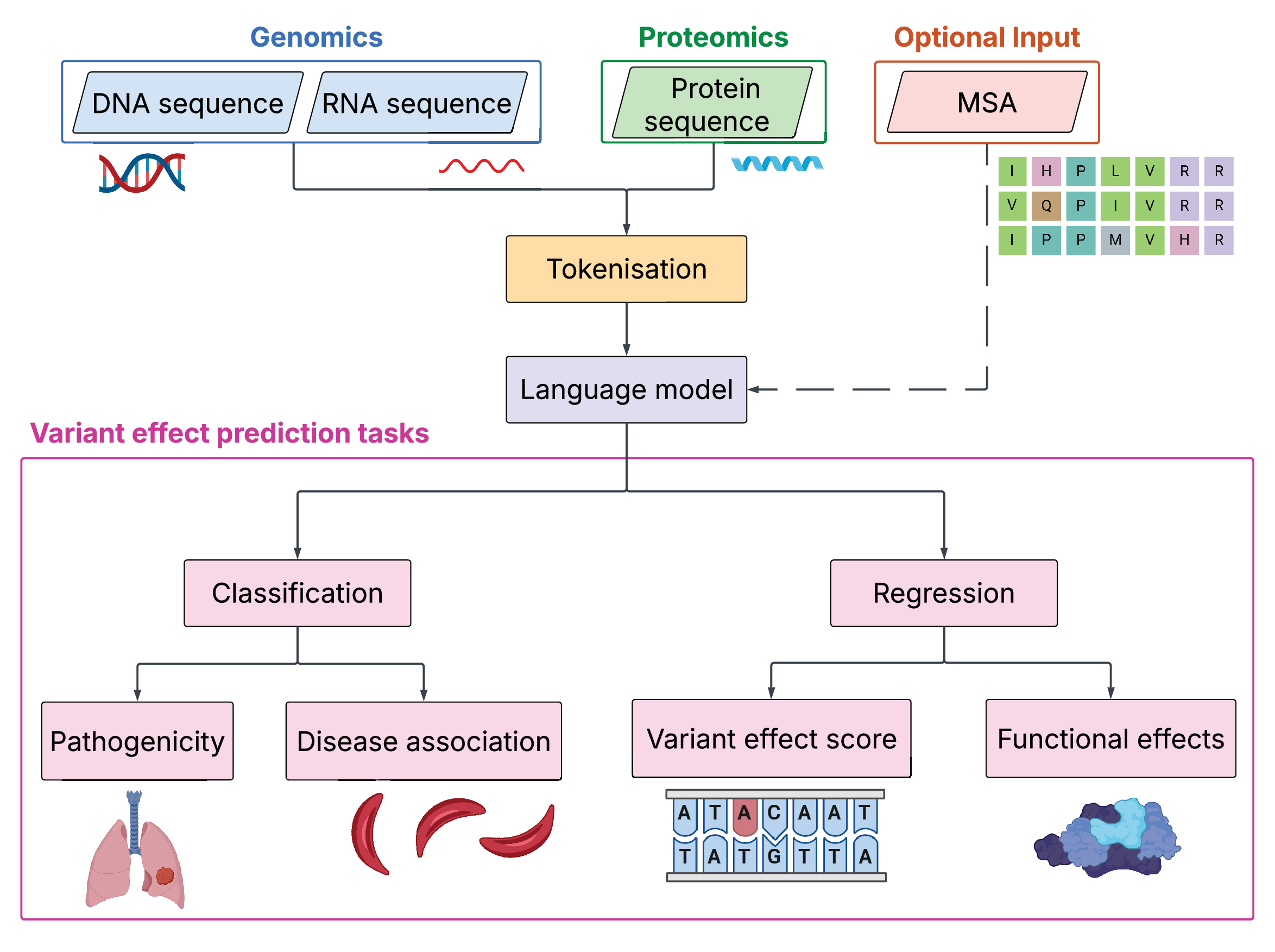}
    \caption{Generic language modelling pipeline, including the main categories of tasks covered in this review. The DNA, RNA, or protein sequences are tokenised before being input to the model. The model is initially pre-trained on a large corpus of data, and then fine-tuned on a dataset specific to the planned downstream tasks, for example, variant pathogenicity classification. Icons from Biorender https://app.biorender.com/.\\
    \textbf{Alt text:} Flowchart showing the language modelling pipeline from inputs to outputs.}
    \label{fig:lmpipeline}
\end{figure}

Natural language processing techniques have long been used to model the structure of DNA, from statistical models \citep{mantegna1995systematic} to large language models \citep{zhang2023applications}. The most frequently observed pipeline among the models reviewed here is shown in Figure \ref{fig:lmpipeline}; the sequences are tokenised before being input to the model, which is first pre-trained on a large corpus of data, and then fine-tuned for specific downstream tasks, such as the examples listed in the figure \citep{ji2021dnabert, tang2023building, diaz2023using}. Although unlabelled datasets of genetic sequences are abundant, labelled datasets are in shorter supply, causing a roadblock in the supervised fine-tuning of LLMs. For variant effect predictors, this can become a concern due to the lack of labelled data related to novel or emerging diseases \citep{nilforoshan2023zero}. However, a small number of models developed in recent years have circumvented the fine-tuning stage by implementing zero-shot prediction \citep{larochelle2008zero}, where models progress straight from pre-training to inference, without needing additional data for fine-tuning \citep{meierlanguage, liu2022protein, nguyen2024evo}.

The first step of the pipeline is tokenisation, where the input sequence is segmented into discrete units, referred to as \textit{tokens}, using defined separators. This process converts the unstructured input data into a standardised format, hence enabling the model to create a numerical representation of the data so it can be processed \citep{webster1992tokenization, palmer2000tokenisation, mielke2021between}. The chosen tokenisation method may have a significant impact on model performance. For instance, \textit{k}-mer tokenisation produces a set of tokens with the same length \textit{k}. The use of these constant-length tokens can lead to heterogeneous token frequencies due to the relative rarity of certain sequence patterns, such as CG dinucleotides \citep{cooper1988cpg}; this can negatively impact the model training process by causing the model to focus on token frequency patterns rather than the contextual relationships within a sequence \citep{sanabria2024dna}. Recent papers have addressed this limitation with the use of byte-pair encoding \citep{gage1994new}, which creates a frequency-balanced vocabulary by creating combined tokens for more frequent sequence patterns \citep{lindsey2024comparison, zhou2024dnabert-2:}.

After tokenisation, the data can be used as an input to the model. There, the selected architecture plays a key role in the quality of predictions produced; the ensuing review will analyse and compare the state-of-the-art architectures in the field.
The concepts of pre-training and fine-tuning date back to the introduction of transfer learning in 1976 \citep{bozinovski1976transfer}. The pre-training stage allows the model to capture knowledge and context that can be used across a wide range of downstream tasks, while the fine-tuning stage builds task-specific understanding \citep{thrun1998learning, pan2009survey}. Pre-training is most frequently done using unsupervised learning tasks such as masked language modelling, on large, unlabelled corpora of genetic sequences; this enables the model to learn without relying on the availability of large labelled datasets, which are scarce in the biomedical field \citep{zhang2023applications, kalyan2022ammu}. The smaller, labelled datasets are then used for task-specific fine-tuning. Frequently used datasets for both stages are detailed in the main review.
After fine-tuning, the model can be used for downstream tasks. Figure \ref{fig:lmpipeline} details some of the most common variant effect prediction tasks. It is important to note that there are several types of variant effect that can be measured, including fitness effect, pathogenicity, and functional change \citep{bromberg2024variant}. These result in different data types, and hence, model functionality will be informed by the specific task at hand. For example, some models may aim to classify a variant as pathogenic or non-pathogenic, whereas others may look to predict a numerical value representing its functional effect \citep{riccio2024variant, bromberg2024variant}.

\section{Language Models for Variant Effect Prediction}
\subsection{Pre-Transformer Models}

\begin{figure}
    \centering
    \includegraphics[width=\linewidth]{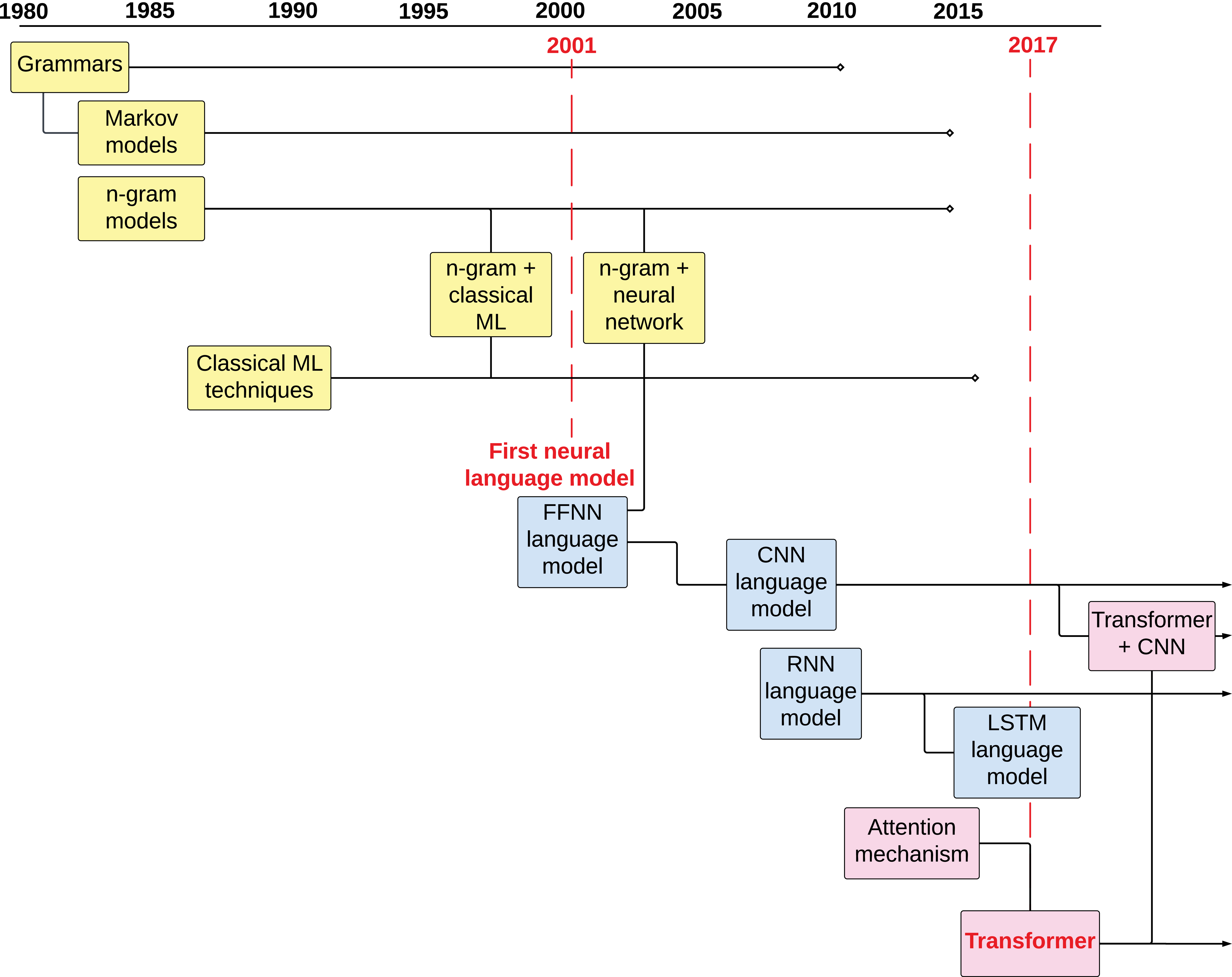}
    \caption{Timeline of models from 1980 until the development of the Transformer. Classical ML refers to classical machine learning techniques such as support vector machine and Naive Bayes. FFNN = feed-forward neural network; CNN = convolutional neural network; LSTM = long short-term memory. Markov models are often used to construct grammars \citep{galley2007lexicalized, zhu2008unsupervised}.\\
    \textbf{Alt text:} Timeline showing the year of emergence of different language modelling techniques, from 1980 to the development of the Transformer in 2017.}
    \label{fig:timeline1}
\end{figure}

Although researchers used forms of language modelling to solve machine translation as early as the 1940s \citep{weaver1949translation, shannon1951redundancy}, Chomsky's work on grammars and syntactical structures in the mid-1950s formed the basis of what we consider natural language processing today, where machines are able to ``understand'' structure and context within languages \citep{chomsky1956three}. A detailed historical review of the field can be found in \citep{jones1994natural}. Since its inception, the field has undergone many changes and innovations; Figure \ref{fig:timeline1} shows the evolution of models up to the development of the Transformer in 2017.

There were significant advancements in the 1980s and 1990s, with the use of statistical models such as n-gram \citep{brown1990statistical} and Hidden Markov Models \citep{ron1996power}. The development of neural networks led to a further turning point in the field, leading to the use of neural language models, which were better able to learn semantic relationships between words, and generalise to unseen test sets, compared to their predecessors \citep{lauriola2022introduction}. The original feed-forward neural network (FFNN) was created in the 1980s \citep{rumelhart1986learning}, and adopted in language modelling in the 2000s \citep{bengio2000neural}. A widely-used neural network architecture is the convolutional neural network (CNN), which was developed in the late 1990s \citep{lecun1998gradientbased}, and introduced in NLP in the mid-2000s \citep{collobert2008aunified}. Instead of relying on manually selected features, CNNs learn features directly from the input data, making them superior to implement end-to-end compared to traditional machine learning methods. Hence, they have become prevalent in DNA sequence modelling and classification \citep{zhou2016cnnsite, kim2017structured, kim2018mut2vec, le2019classifying}. While CNNs are excellent at learning short-range dependencies, they struggle to model relationships between words (or nucleotides) far from each other \citep{wang2018application}. This limitation underscores the need for more advanced architectures to address such dependencies.

Recurrent neural networks (RNNs) \citep{rumelhart1986learning} were introduced in NLP as a possible alternative to CNNs, as the use of recurrent connections enabled these models to incorporate many previous inputs into future steps \citep{mikolov2010rnn}. However, traditional RNNs suffer from a problem referred to as ``vanishing gradients'', which makes them prone to ``forgetting'' inputs that are further back in the sequence. Two main alternatives have been brought forth in an attempt to circumvent this problem: (i) the long short-term memory network (LSTM) \citep{hochreiter1997long}, which is able to handle long-term dependencies using a more complex architecture formed of different gates, and (ii) the gated recurrent unit (GRU) \citep{cho2014learning}, which uses a simplified version of the LSTM architecture to streamline sequence handling. Several variants of these models have been utilised for language modelling over the past decades, both individually and as part of ensembles with other neural networks such as CNNs \citep{rhanoui2019cnnbilstm, lin2020crispr}.

A significant limitation common to statistical and neural language models is the need to specify a fixed context length prior to training; this restricts the capacity of these models to utilise extended contexts for predictions \citep{liu2024lost}. The attention mechanism was created to address this limitation, by computing weights for each token in the input sequence to capture its relation to the others, and applying scaling to focus (or ``give attention'') on the tokens relevant to the task \citep{bahdanau2014neural}. Several models achieved good results on machine translation tasks by combining this attention mechanism with recurrent networks \citep{wu2016google, kim2017structured}.
The attention mechanism was eventually developed into the self-attention mechanism, which forms the basis of the modern Transformer \citep{vaswani2017attentionisallyouneed}. Self-attention (Figure \ref{fig:attention_and_alternatives}) is applied within a single sequence to compute a representation of that sequence, and provides a method of learning long-range dependencies within input sequences. The Transformer architecture combines self-attention with fully-connected layers, which are stacked to create an encoder-decoder model. Multiple self-attention mechanisms are used in parallel; this is referred to as multi-head self-attention (Figure \ref{fig:attention_and_alternatives}), and reduces the complexity per layer, hence increasing the capability for parallelisation. These features of the Transformer make it more performant on complex tasks compared to recurrent or convolutional networks, and also increase its efficiency. As illustrated in Figure \ref{fig:timeline1}, the Transformer has been used both independently, and in conjunction with other models such as LSTM and CNN.

Despite the fact that the introduction of Transformers in 2017 marked a significant milestone in deep learning, the development of models using other architectures has continued. As shown in Table \ref{tab:neural}, many recent models using pre-Transformer technologies, including those using CNNs, have demonstrated notable performance.
In particular, the GPN model, a CNN-based approach for genome-wide effects of variants in DNA, has demonstrated state-of-the-art performance \citep{benegasdna}. The architecture of the convolutional model was selected after it was observed that it converged faster than its Transformer-based counterpart during pre-training, and the results showed that it outperformed other genome-wide variant effect predictors for \textit{Arabidopsis}. Another noteworthy finding of this study was the performance gain observed from training on multi-species data instead of single-species data. This suggests that incorporating cross-species data can provide richer context for understanding genetic variation, and can potentially improve the generalisability of the model.

In addition to CNNs, the graph convolutional network (GCN) has also proved to be a performant non-Transformer language modelling approach for variant effect prediction. Notably, its enhanced ability to capture graph-like structural information compared to other neural network architectures has proven useful in DNA variant effect prediction approaches incorporating structural data alongside sequence data \citep{tan2023multimodal}.

These findings underscore the ongoing relevance of pre-Transformer neural network architectures in genomics, and highlight the potential benefits of leveraging diverse datasets for training.

\begin{table*}[t]
\caption{Summary of neural language models for variant effect prediction. See Table \ref{supp:neural} for code/data availability.\label{tab:neural}}
\tabcolsep=0pt
\begin{tabular*}{\textwidth}{@{\extracolsep{\fill}}lp{0.3\textwidth}llll@{\extracolsep{\fill}}}
\toprule
Paper & Task & Year & Architecture & Data Type & Variant Type\\
\midrule
    \citep{kim2018mut2vec} & Identifying cancer driver mutations & 2018 & CNN & DNA & Coding\\ 
    \citep{pejaver2020inferring} & Inferring the molecular and phenotypic impact of SAVs & 2020 & CNN & Protein & Coding\\
    \citep{shin2021protein} & Protein variant effect prediction & 2021 & CNN & Protein, RNA & \\
    \citep{dunham2023high-throughput} & Protein variant effect prediction & 2023 & CNN & Protein & Coding\\
    \citep{benegasdna} & Prediction of genome-wide DNA variant effects & 2023 & CNN & DNA & Coding, Non-coding\\
    \citep{tan2023multimodal} & Non-coding variant effect prediction using genome sequence and chromatin structure & 2023 & CNN, GCN & DNA & Non-coding\\
    \citep{cheng2023self-supervised} & Self-supervised Learning for DNA sequences with circular dilated convolutional networks & 2024 & CNN & DNA & Non-coding\\
\bottomrule
\end{tabular*}
\end{table*}

\subsection{Transformer-Based Models}

\subsubsection{History \& Overview}

\begin{figure}[!t]
\centering
\includegraphics[width=\linewidth]{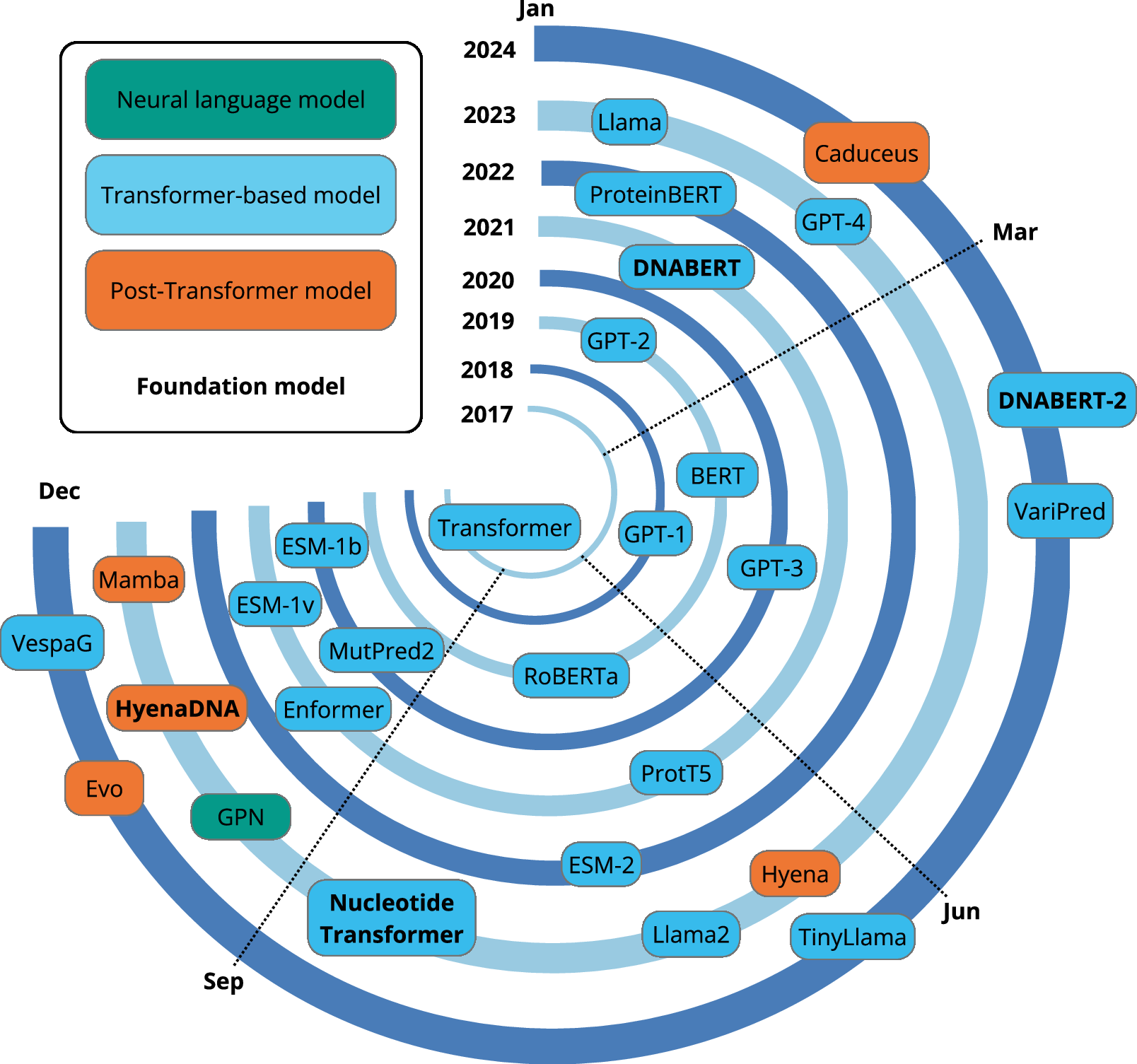}
\caption{Timeline of developments in NLP since 2017.\\
\textbf{Alt text:} Radial timeline showing the years in which impactful language modelling technologies were developed, starting with the Transformer in 2017.}\label{timeline2}
\end{figure}

\begin{figure}[!t]
    \centering
    \includegraphics[width=\linewidth]{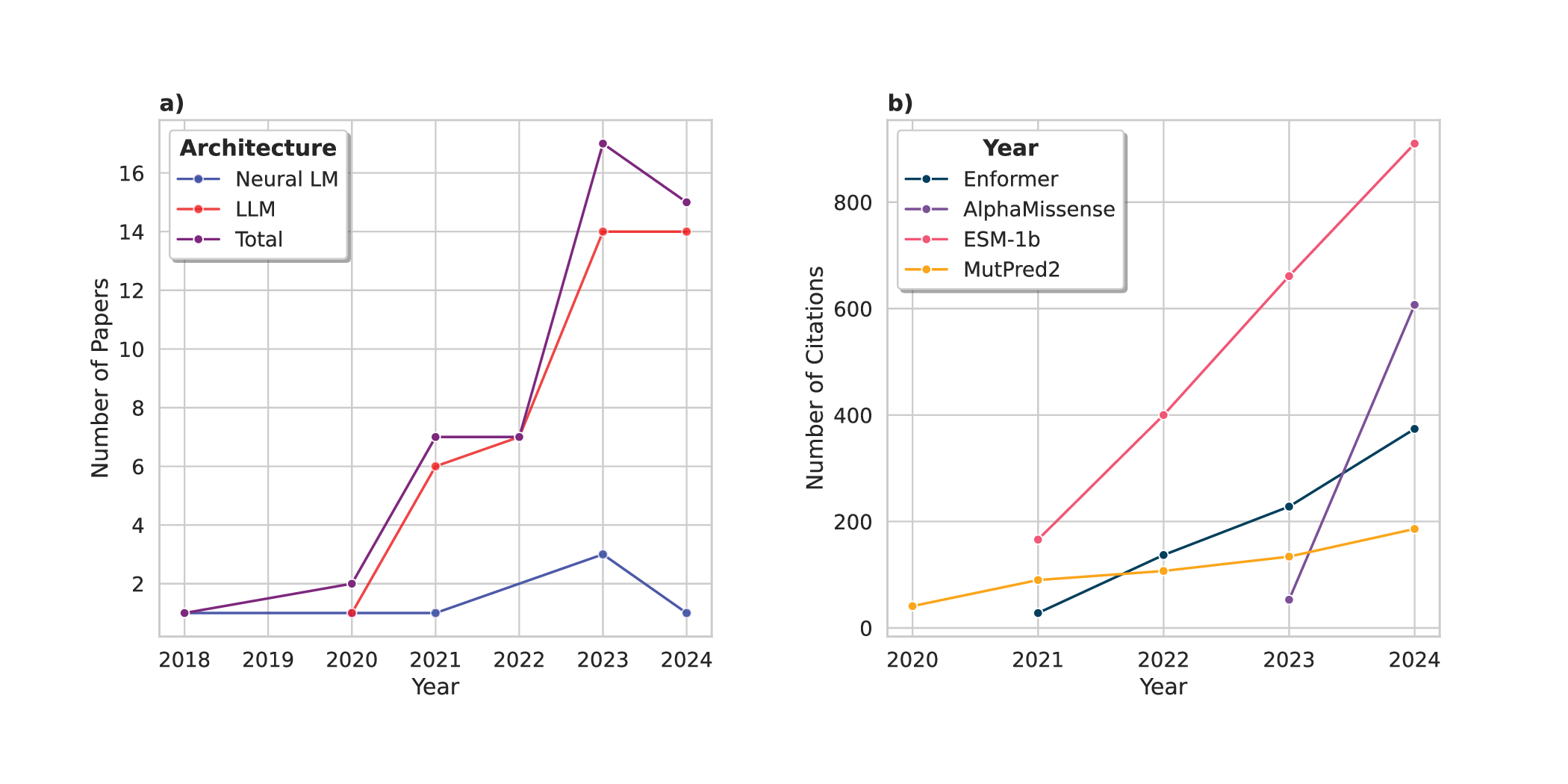}
    \caption{Analysis of the number of published papers, and the number of annual citations for the highest-impact papers. \textbf{(a)} Number of papers published per year on language models for variant effect prediction, as described in Tables \ref{tab:neural}, \ref{tab:transformer_models}, and \ref{tab:post_transformer_models}. Neural LM = neural language models (Table \ref{tab:neural}). LLM refers to both Transformer-based and post-Transformer models (Tables \ref{tab:transformer_models} and \ref{tab:post_transformer_models}). During the period 2018-24, the overall number of papers per year has generally increased, with a slight decrease from 2023 to 2024.  The number of LLM papers has far exceeded the number of neural LM papers each year. \textbf{(b)} Number of citations per year for the most impactful papers. The number of citations per year for these papers has steadily increased since their publication.\\
    \textbf{Alt text:} Graphs on paper publication and citation data with sub-figures labelled a and b.}
    \label{fig:num_models}
\end{figure}

\begin{figure}[!t]
\centering
\includegraphics[width=\linewidth]{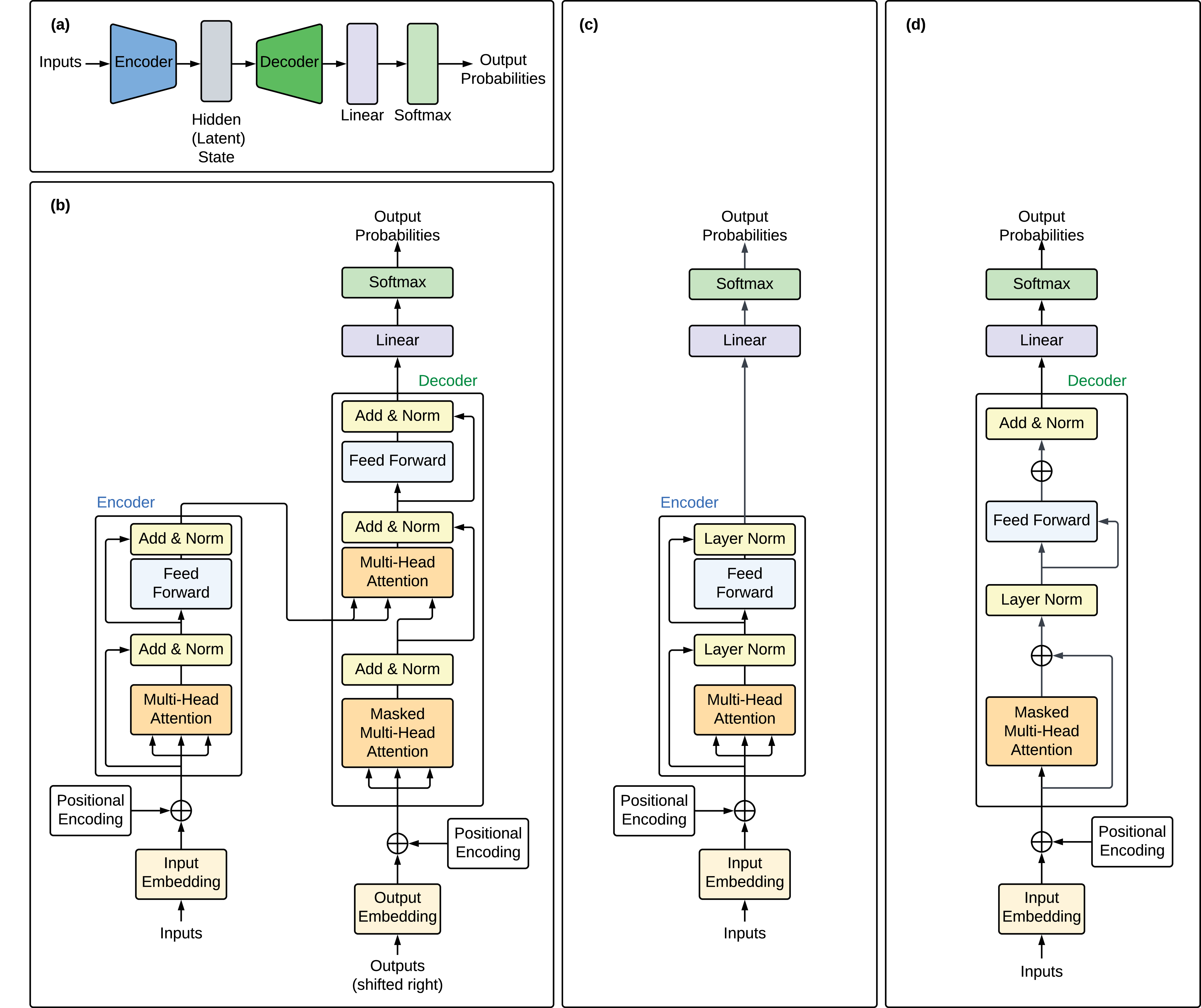}
\caption{Transformer architectures. (a) High level representation of the encoder-decoder architecture comprising the vanilla Transformer architecture. The encoder encodes the input sequence into a representation, which is stored as a latent state. The decoder decodes this representation into an output sequence. This is passed into the linear and softmax layers to produce the output predictions. (b) Detailed Transformer architecture, adapted from \citep{vaswani2017attentionisallyouneed}. The multi-head attention modules consist of multiple self-attention modules used in parallel. These are stacked with fully-connected layers to create an encoder-decoder model as shown in (a). (c) Encoder-only Transformer architecture, adapted from DNABERT \citep{ji2021dnabert}. (d) Decoder-only Transformer architecture, adapted from GPT-1 \citep{radford2018improving}.\\
\textbf{Alt text:} Diagrams of Transformer architectures, with sub-figures labelled from a to d.}\label{fig:llm_architectures}
\end{figure}

The origination of the Transformer architecture was a pivotal point in the natural language processing field, resulting in models that pushed the boundaries of human ability to process natural and biological languages. Figure \ref{timeline2} summarises the timeline of the most impactful models that have been produced, starting with the original Transformer in 2017. After seven years, it is still an active field of investigation; 2023, in particular, was a year of many developments for both Transformer-based and non-Transformer language models.
The original Transformer architecture is summarised in Figure \ref{fig:llm_architectures}(a) and shown in detail in Figure \ref{fig:llm_architectures}(b) \citep{vaswani2017attentionisallyouneed}. The multi-head attention modules consist of multiple self-attention modules used in parallel. These are stacked with fully-connected layers to form an encoder-decoder model. The input sequence is encoded by the encoder into a representation which is then stored as a latent state. The decoder then decodes the representation into an output sequence, which is subsequently passed to the linear and softmax layers to produce the output predictions. While the original Transformer uses an encoder-decoder architecture, it is possible to have models consisting of only one or the other.

For instance, the GPT series of models \citep{radford2018improving, floridi2020gpt, achiam2023gpt} are decoder-only generative models, which, when given an input sequence, output the probabilities of possible subsequent tokens. By feeding the extended sequence back into the model, and repeating the process many times, it is possible to generate a body of text. Figure \ref{fig:llm_architectures}(d) shows a decoder-only model based on GPT-1 \citep{radford2018improving}. These models have undergone significant developments since the release of GPT-1 \citep{radford2018improving}, and now form the basis of the notorious ChatGPT chatbot.
A significant limitation of models using the standard Transformer architecture is their unidirectionality; each token can only incorporate context from the previous tokens, hence limiting the model's ability to perform sentence-level tasks \citep{devlin2019bertpretrainingdeepbidirectional, shreyashree2022literature}. This was addressed by the development of BERT (Bidirectional Encoder Representations from Transformers) \citep{devlin2019bertpretrainingdeepbidirectional}, an encoder-only model which transforms text embeddings into a representation that can be used for a variety of tasks. BERT achieves bidirectionality by using a masked language modelling (MLM) pre-training objective, in which the model attempts to predict the identity of randomly masked tokens in the input sequences, hence learning a representation that combines the context from the left and right. Although originally designed to process text, BERT has also been extensively applied in the field of molecular biology, resulting in models such as DNABERT \citep{ji2021dnabert} (Figure \ref{fig:llm_architectures}(c)) and ProteinBERT \citep{brandes2022proteinbert}.
Though Transformer-based models have been most commonly used in the field, CNNs remain in use, both as the basis of models such as GPN \citep{benegasdna}, and in conjunction with Transformers in models such as Enformer \citep{avsec2021effective}.

Though LLMs have led to a paradigm shift in computational solutions for biological problems, they still experience several limitations. Data scarcity is a significant challenge; limited high-quality labelled data is available for several biological problems of interest, including non-coding variant effect prediction \citep{wang2024deep, yang2024novo}. This limits the use of LLMs for these problems due to their requirement for large quantities of training data. Additionally, training on insufficiently diverse data can lead to poor generalisation across tasks \citep{zhang2023applications}.
Efforts to address these limitations have led to the emergence of foundation models, LLMs which are pre-trained on very large-scale data for parameter initialisation and are then able to be fine-tuned for an extensive range of downstream applications \citep{bommasani2022opportunities, zhou2024comprehensive}. The data-intensive pre-training stage enables fine-tuning with comparatively limited data, hence improving the models' generalisability and allowing the models to be applied to biological problems with insufficient data to train an LLM from scratch \citep{li2024progress}. Notable foundation models in bioinformatics, highlighted in red text on Figure \ref{timeline2}, are DNABERT \citep{ji2021dnabert}, DNABERT-2 \citep{zhou2024dnabert-2:}, Nucleotide Transformer \citep{dalla2024nucleotide}, and the ESM series of models \citep{rives2021biological, meierlanguage, lin2023evolutionary}.

Despite the many successes of Transformers, they also have a major drawback: the time and memory used by the self-attention mechanism scale quadratically with sequence length, leading to high computational costs and creating a performance bottleneck \citep{zhu2021longshorttransformer, poli2023hyena, consens2023transformersbeyondlargelanguage}. These models are hence impractical to train and use without access to extensive computational equipment and power. Crucially, this is also an environmental concern, with LLMs having huge carbon and water footprints \citep{huang2024white, jiang2024preventing}. Hence, research is required to produce models that can achieve excellent results without being highly resource-instensive. These concerns have sparked a trend in the field of creating computationally efficient models as an alternative to the Transformer; these are explored in detail in the next section.
Notwithstanding the benefits of these post-Transformer technologies, development of Transformer-based models has continued, with the release of highly-performant models such as DNABERT-2 \citep{zhou2024dnabert-2:} and VespaG \citep{marquet2024vespag:} as recently as 2024.

\subsubsection{Review of Existing Models}
Transformer-based LLMs are by far the most common language models used in the variant effect prediction field. This section reviews the existing models in the field, identifying key trends.

While all models surveyed take a sequence input - DNA, protein, or RNA - the precise input type varies. Some models take both the mutated and wild-type sequences as input \citep{derbel2023accurate, lin2024enhancing, zhan2024dyna:, yan2024transefvp:}, while others take a wild-type sequence alongside tabular data describing a variant \citep{pejaver2020inferring, manfredi2022e-snpsgo:}. Whereas the majority of models report taking an input sequence of length up to 10,000 bases (Figure \ref{fig:len_params_time}), the Enformer \citep{avsec2021effective} is notable as it can process significantly longer sequences, i.e., up to 96,608 bases.

In addition to sequence input, several methods integrate multiple sequence alignments (MSA) as an additional input. Indeed, the conserved residues predicted by MSA can be predictive of variant effect \citep{liu2014quantitative, capriotti2022evaluating}. Thus, it has been observed across many models that incorporating MSA as an auxiliary form of data improves the quality of predictions \citep{blaabjerg2024ssemb, benegas2023gpn-msa}. However, this is largely dependent on the quality of the MSA, which is variable, and often poor due to a lack of appropriate data \citep{livesey2020using, ranwez2020strengths}.
Despite the positive results observed in variant effect predictors using MSA, they are not appropriate for all use cases, as many variants lie outside MSA coverage \citep{brandes2023genome-wide}. Additionally, several predictors not using MSA have matched or outperformed MSA-based predictors while eliminating the additional computational cost associated with having a larger training dataset \citep{lin2024enhancing}. For example, a benchmarking study \citep{livesey2023updated} showed that ESM-1v \citep{meierlanguage}, which does not use MSA, outperformed several MSA-based state-of-the-art models. Hence, many recent approaches to variant effect prediction have eschewed MSA in favour of sequence-only input.

Human data is most predominantly used to train and test the models surveyed here. However, a few studies have demonstrated that incorporating data from multiple species during training can improve results compared to models trained on human data only. Indeed, it has been suggested that learning the variability across various genomes can assist a model in learning about the degree of conservation across genetic sites, hence improving its ability to predict variant pathogenicity \citep{rao2021msa, dalla2024nucleotide, benegas2023gpn-msa}.

The majority of models surveyed adhere to the pipeline described in Figure \ref{fig:lmpipeline}, which includes pre-training and fine-tuning stages. Traditionally, language models used the pre-training task of next-token prediction. While this is still used in some contemporary models \citep{poli2023hyena}, the field has generally moved to favour masked language modelling (MLM) \citep{devlin2019bertpretrainingdeepbidirectional} due to its ability to incorporate bidirectional context. However, MLM is not always the optimal choice, as it has been suggested that it may be insufficiently challenging for the model in cases where the training data includes a multi-species MSA containing sequences very similar to the human genome; this has previously been addressed by excluding these very similar genomes during training \citep{benegas2023gpn-msa}.

To maximise efficiency and minimise computational cost, recent work has explored zero-shot prediction, where prediction is performed straight after pre-training, without fine-tuning. A benchmarking study \citep{tang2023building} compared the ability of several state-of-the-art models to perform a non-coding variant effect prediction task \citep{shigaki2019integration} without additional fine-tuning. There, two Transformer models, i.e., Nucleotide Transformer \citep{dalla2024nucleotide} and Enformer \citep{avsec2021effective}, were compared with CNN models GPN \citep{benegasdna} and ResidualBind \citep{koo2021global}. Eventually, Enformer performed best, achieving a Pearson correlation of 0.68 between the experimental and predicted values. Then, the CNN methods achieved correlations between 0.35-0.55, whereas Nucleotide Transformer performed worst, with a correlation lower than 0.1. Based on these results, it was suggested that specialised supervised models may be a better choice for zero-shot prediction compared to current LLMs, which are pre-trained on broad datasets \citep{tang2023building}.

While the original Transformer architecture consists of an encoder-decoder framework (Figure \ref{fig:llm_architectures}(a),(b)), the decoder portion is often not required for biological language models, as sequence generation tasks are uncommon in this field. Hence, the majority of models summarised in Table \ref{tab:transformer_models} employ an encoder-only framework, often based on BERT to implement bidirectionality (Figure \ref{fig:llm_architectures}(c)). Indeed, state-of-the-art papers have demonstrated that such architectures are able to successfully model genetic sequences without the need for a decoder \citep{ji2021dnabert, zhou2024dnabert-2:, rives2021biological, meierlanguage}. Still, a few encoder-decoder models, based on the original Transformer \citep{vaswani2017attentionisallyouneed}, are also present \citep{avsec2021effective, hidayat2023utilizing, truongpoet:, gao2023epigept:}. There is a lack of decoder-only models, however, this is to be expected, as such models are generally better-suited to generating sequences, an ability which is not required for most variant effect prediction tasks.
Furthermore, novelty does not always reside in the architecture; many models are based on pre-trained LLMs, which are then fine-tuned, hence eliminating the additional time and computational expense associated with pre-training a new model for a similar set of tasks. A prominent example is ESM-1b \citep{rives2021biological}, which has been exploited by many studies attempting protein variant effect prediction, as shown in Table \ref{tab:transformer_models}.
Another use of pre-trained models in the field has been to provide input into models that can be considered meta-predictors \citep{liu2022computational}. Such models input data into a pre-trained LLM, extract the output embeddings, and add a simple neural network based classifier or regressor on top to make predictions based on these embeddings. This approach is highly data- and time-efficient in comparison to other LLM workflows, as it eliminates any training or fine-tuning of the LLM, and requires only training of a simple neural network. Models using this methodology have achieved state-of-the-art results, showcasing this as an accurate and efficient framework for variant effect prediction \citep{avsec2021effective, wild2024dna}. Recent work has also discovered benefits from integrating embeddings from multiple pre-trained LLMs, hence combining important context from diverse sources \citep{yan2024transefvp:}.

\begin{figure}
    \centering
    \includegraphics[width=\linewidth]{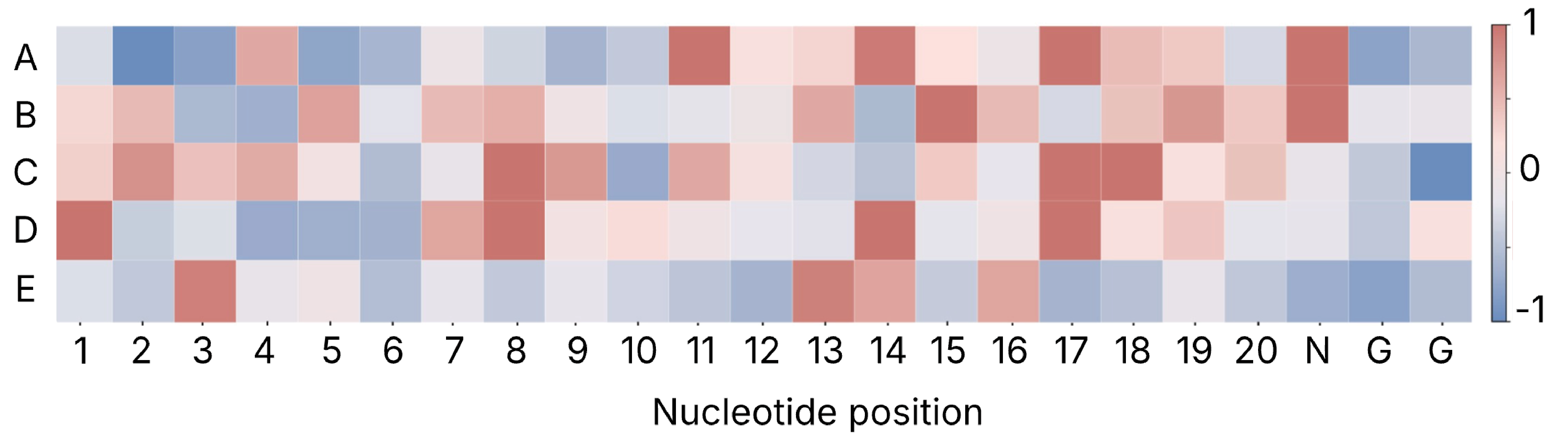}
    \caption{Heatmap adapted from that produced in \citep{luo2024interpretable} using the Deep SHAP method \citep{scott2017unified}. Evaluation was done on five independent datasets, each for a different cell line. The y-axis denotes the dataset, while the x-axis denotes the nucleotide position. The colours indicate the importance of the nucleotide position towards the predicted class label; the legend is shown on the right-hand side. 1 and -1 respectively indicate a significant positive or negative contribution. \textbf{Note:} A key element of the CRISPR-Cas9 DNA editing system is the single-guide RNA sequence consisting of a 20-nucleotide protospacer, and a 3-nucleotide protospacer adjacent motif (PAM) sequence \citep{fuguo2017crispr}. The "N", "G", and "G" positions represent the PAM sequence, which consists of any one nucleotide (N) followed by two guanines (GG).\\
    \textbf{Alt text:} Image of a heatmap with coloured squares to indicate the strength of the relationship between the x and y axes.}
    \label{fig:luo2024interpretable_heatmap}
\end{figure}

While significant developments in model architecture have occurred, work on model interpretability is still limited. The majority of models mentioned in Table \ref{tab:transformer_models} function as black boxes, taking an input, and returning an output. Although some of them have provided promising results, it is difficult for humans to understand and interpret the underlying logic. Currently, it is uncommon for this issue to be addressed in papers in the field; however, a recent study on predicting CRISPR/Cas9 off-target activities included interpretability as a key contribution \citep{luo2024interpretable}. In the CRISPR/Cas9 gene editing system, base mismatches can occur during pairing of DNA and single-guide RNA sequences, leading to poor gene editing outcomes, and increasing the risk of ``off-target'' mutations. Deep SHAP \citep{scott2017unified}, a statistical method to calculate the contribution of each hidden unit to the predictions of a model, was used to evaluate the importance of specific nucleotide positions in the model's classification of off-target or on-target for each single-guide RNA and DNA pair. This method is easily interpretable by humans, and can be used to plot a heatmap to visually identify key positions which contribute significantly to the decision-making process of the model. The resultant heatmap from the paper is shown in Figure \ref{fig:luo2024interpretable_heatmap} \citep{luo2024interpretable}. The colour of each square indicates the strength of the contribution of the nucleotide position to the predicted class label; the legend is shown on the right-hand side.

The developments described above have resulted in the models described in Table \ref{tab:transformer_models}. Comparing the performance of models across papers is challenging, as different studies tend to evaluate models on different datasets often using different metrics. One therefore cannot definitively conclude that a certain model is state-of-the-art in all aspects. It is, however, possible to assess trends across the models for specific tasks. For instance, Transformer-based models have demonstrated good performance in classifying single amino acid variant (SAV) pathogenicity from protein sequences, with a number of models achieving AUROC $>$ 0.8 \citep{manfredi2022e-snpsgo:, jiang2023deciphering, sun2023structure-informed, lafita2024machine}, and a few studies achieving AUROC $>$ 0.9 \citep{rives2021biological, brandes2022proteinbert, james2023deep}. The unique study published on predicting the effects of protein indels also showed promising performance; AUROC $>$ 0.8 was achieved when predicting the pathogenicity of both insertions and deletions across two separate datasets. However, this outperformed the previous state-of-the-art (non-Transformer) methods by less than 0.1.
Outcomes in predicting the functional scores of protein variants show a greater degree of variability, with the correlation between true and predicted values varying from below 0.5 \citep{blaabjerg2024ssemb} to above 0.9 \citep{rives2021biological}. However, this significant disparity in results may be due to the fact that these models were evaluated on different datasets.
Performance on DNA variant effect prediction is similarly varied, though the best-performing models have achieved AUROC $>$0.9 for SNP classification \citep{yang2022integrating, benegas2023gpn-msa, gao2023epigept:, zhan2024dyna:}. Although both coding and non-coding regions are addressed by these models, performance on some non-coding variant effect prediction tasks is still low; for instance, state-of-the-art models have achieved a correlation of less than 0.6 between true and predicted values on the Variant Effect Causal eQTL dataset (Table \ref{tab:datasets}) \citep{avsec2021effective}.
Existing work on RNA tasks is promising, though limited. The evaluation of three models on a SARS-CoV-2 variant classification task yielded a best F1-score of 73.04, indicating potential for further enhancement \citep{zhou2024dnabert-2:}.
Overall, the models demonstrating state-of-the-art performance across multiple tasks have been the Nucleotide Transformer \citep{dalla2024nucleotide}, DNABERT-2 \citep{zhou2024dnabert-2:}, and ESM-1b \citep{rives2021biological}. These are all foundation models, the former two for DNA, and the latter for proteins. These results suggest that foundation models represent a promising direction for future research.

The Transformer has led to a plethora of interesting and valuable studies on variant effect prediction. However, the lack of standard evaluation datasets and protocols has made performance comparison particularly difficult. Overall, performance on protein variant pathogenicity classification has been high, however, non-coding DNA and RNA variant effect prediction tasks have proved challenging and, thus, require further investigation to improve results. Recent approaches have aimed to reduce the computational cost associated with training and testing Transformer-based models alongside enhancing the prediction quality.
The increasing number of papers published on such models since 2020 (Figure \ref{fig:num_models}), and the fact that such papers have been published as recently as January 2025 (Table \ref{tab:transformer_models}), suggests that the Transformer remains competitive for variant effect prediction.

\begin{landscape}

\begin{longtable}{lp{0.4\linewidth}p{0.05\linewidth}p{0.2\linewidth}ll}
\caption{Summary of Transformer-based language models for variant effect prediction. * = preprint.} \label{tab:transformer_models}

\\\hline \multicolumn{1}{c}{Paper} & \multicolumn{1}{c}{Task} & \multicolumn{1}{c}{Year} & \multicolumn{1}{c}{Architecture} & \multicolumn{1}{c}{Data Type} & \multicolumn{1}{c}{Variant Type} \\ \hline 
\endfirsthead

\multicolumn{6}{c}%
{{\bfseries \tablename\ \thetable{} -- continued from previous page}} \\
\hline \multicolumn{1}{c}{Paper} & \multicolumn{1}{c}{Task} & \multicolumn{1}{c}{Year} & \multicolumn{1}{c}{Architecture} & \multicolumn{1}{c}{Data Type} & \multicolumn{1}{c}{Variant Type} \\ \hline 
\endhead

\hline \multicolumn{6}{r}{{Continued on next page}} \\ \hline
\endfoot

\hline \hline
\endlastfoot
   
    \citep{li2020predicting} & Prediction of pathogenicity of protein sequences & 2020 & Encoder-only (BERT) & Protein & Coding\\
    \citep{rives2021biological} & Prediction of protein variant effects & 2020 & ESM-1b - Encoder-only & Protein & \\
    \citep{meierlanguage} & Prediction of functional effects of protein mutations & 2021 & ESM-1v - Encoder-only & Protein & \\
    \citep{amadeus2021design} & Polygenic risk model for colorectal cancer & 2021 & Encoder-only & DNA & Coding, Non-coding\\
    \citep{avsec2021effective} & Prediction of non-coding DNA variants effects on gene expression & 2021 & Encoder-decoder & DNA & Non-coding\\
    \citep{ji2021dnabert} & Identification of functional variants in non-coding DNA & 2021 & Encoder-only & DNA & Non-coding\\
    \citep{yamaguchi2021evotuning} & Prediction of variant effects on multi-domain proteins & 2021 & Encoder-only & Protein & \\
    \citep{liu2022protein}* & Zero-shot protein mutation pathogenicity prediction & 2022 & ESM-1b \citep{rives2021biological} & Protein, MSA & \\
    \citep{marquet2021embeddings} & Prediction of protein variant effects & 2022 & ProtBert \citep{elnaggar2021prottrans}, ESM-1b \citep{rives2021biological}, ProtT5-XL-U50 \citep{elnaggar2021prottrans} & Protein & \\
    \citep{yang2022integrating} & Prediction of deleteriousness of SNPs in non-coding DNA & 2022 & Encoder-only & DNA & Non-coding\\
    \citep{olenyi2022lambdapp:} & Predicting SAV effects & 2022 & \citep{marquet2021embeddings} & Protein & \\
    \citep{zhou2022unsupervised}* & Prediction of protein variant pathogenicity & 2022 & \citep{elnaggar2021prottrans} & Protein & \\
    \citep{manfredi2022e-snpsgo:} & Prediction of SAV pathogenicity & 2022 & ESM-1v \citep{meierlanguage}, \citep{elnaggar2021prottrans} & Protein & Coding \\
    \citep{dampier2022hiv} & Prediction of protease inhibitor resistance in HIV-1 mutations & 2022 & Encoder-only (BERT) & RNA & \\
    \citep{jiang2023deciphering} & Prediction of SAV pathogenicity & 2023 & Encoder-only (BERT) & Protein & \\
    \citep{sun2023structure-informed}* & Prediction of SAV pathogenicity from sequence and structure & 2023 & Encoder-only (BERT) & Protein, MSA & \\
    \citep{brandes2023genome-wide} & Prediction of protein variant pathogenicity & 2023 & ESM-1b \citep{rives2021biological} & Protein & \\
    \citep{fan2022shine:} & Prediction of pathogenicity of insertion and deletion variants from protein sequences & 2023 & ESM-1b \citep{rives2021biological}, \citep{rao2021msa} & Protein, MSA & \\
    \citep{benegas2023gpn-msa}* & Genome-wide variant effect prediction in human DNA & 2023 & Encoder-only & DNA, MSA & Coding, Non-coding\\
    \citep{derbel2023accurate} & Prediction of functional effect of SAVs & 2023 &  & Protein & Coding\\
    \citep{hidayat2023utilizing} & Prediction of BRCA1 variant pathogenicity & 2023 & ESM2 \citep{lin2023evolutionary} & DNA & Coding\\
    \citep{james2023deep} & Prediction of protein-coding SAV pathogenicity in the low density lipoprotein receptor (LDLR) protein & 2023 & \citep{frazer2021disease}, ESM-1v \citep{meierlanguage}, \citep{jumper2021highly} & Protein, MSA & Coding\\
    \citep{zhou2024dnabert-2:}* & SARS-CoV-2 variant classification & 2023 & Encoder-only & RNA & Non-coding\\
    \citep{cheng2023accurate} & Proteome-wide missense variant effect prediction & 2023 & Encoder-only, based on \citep{jumper2021highly} & Protein & \\
    \citep{danzi2023deep} & Variant prioritisation in Mendelian diseases & 2023 & \citep{elnaggar2021prottrans} & Protein, MSA & Coding\\
    \citep{truongpoet:} & Prediction of protein variant fitness & 2023 & Encoder-Decoder & Protein, MSA & \\
    \citep{qu2023ensemble} & Prediction of protein mutation effects using ensemble learning & 2023 & Ensemble: \citep{notin2022tranception}, \citep{vaswani2017attentionisallyouneed} & Protein, MSA & \\
    \citep{dalla2024nucleotide} & Prediction of DNA variant effects & 2024 & Encoder-only & DNA & Coding, Non-coding\\
    \citep{blaabjerg2024ssemb} & Protein variant effect prediction from sequence and structure & 2023 & Based on \citep{rao2021msa} & Protein, MSA & \\
    \citep{lin2024enhancing} & Prediction of protein missense variant pathogenicity & 2024 & ESM-1b \citep{rives2021biological} used in twin network configuration & Protein & Coding\\
    \citep{wild2024dna}* & Prediction of DNA variant pathogenicity & 2024 & \citep{benegas2023gpn-msa}, \citep{dalla2024nucleotide} & DNA & Coding, Non-coding\\
    \citep{luo2024interpretable} & Prediction of off-target effects of mismatches and indels & 2024 & Encoder-only (BERT) & DNA, RNA & \\
    \citep{gao2023epigept:}* & (1) Prediction of DNA variant effects (2) SARS-CoV-2 variant prioritisation & 2024 & Encoder-decoder & DNA, RNA & Non-coding\\
    \citep{zhan2024dyna:}* & Prediction of coding and non-coding variant effects & 2024 & ESM-1b \citep{rives2021biological} & DNA, Protein & Coding, Non-coding\\
    \citep{lafita2024machine}* & Prediction of SAV pathogenicity & 2024 & ESM-1b \citep{rives2021biological}, ESM-1v \citep{meierlanguage}, ESM2 \citep{lin2023evolutionary} & Protein & Coding\\
    \citep{marquet2024vespag:} & Prediction of SAV effect score & 2024 & Shallow neural network on top of \citep{lin2023evolutionary} & Protein & \\
    \citep{yan2024transefvp:} & Prediction of SAV pathogenicity & 2024 & Ensemble: ESM-1b \citep{rives2021biological}, ESM-1v \citep{meierlanguage}, ESM2 \citep{lin2023evolutionary}, \citep{elnaggar2021prottrans} & Protein & Coding\\
    \citep{Shulgina2024.04.05.588317}* & Identification of RNA mutations beneficial to thermostability & 2024 & Decoder-only (GPT) & RNA & \\
    \citep{yang2024transvpath} & Pathogenicity scoring for structural variants & 2024 & TabTransformer \citep{huang2020tabtransformer} & DNA & Coding\\
    \citep{li2024mvformer} & Prediction of missense coding variant pathogenicity & 2024 & Gated Transformer & Protein & Coding\\
    \citep{zhong2024premode}* & Prediction of functional effects of protein missense variants & Graph attention Transformer & 2024 & Protein & Coding\\
    \citep{linder2025predicting} & Predicting the impact of genetic variation on gene expression & 2023 & Encoder-Decoder (based on \citep{avsec2021effective}) & DNA & Coding, Non-coding\\
    \citep{joshi2025augmented} & Prediction of coding VUS pathogenicity & 2025 & ESM-1b & DNA & Coding\\
    \citep{glaser2025esm}* & Prediction of functional effect of protein mutations & 2025 & ESM2 \citep{lin2023evolutionary} & Protein & Coding\\
\end{longtable}

\begin{table*}[ht!]
\caption{Summary of existing benchmarks for large language models in variant effect prediction field. See Table \ref{supp:benchmarks} for access links.\label{tab:benchmarks}}
\tabcolsep=2pt
\begin{tabular*}{\linewidth}{@{\extracolsep{\fill}}p{0.3\textwidth}p{0.25\textwidth}lllll@{\extracolsep{\fill}}}
\toprule
Benchmark & Task & Year & Data Type & No. Samples & Organisms & No. Predictors Evaluated\\
\midrule
    Benchmarking of variant effect predictors using deep mutational scanning \citep{livesey2020using} & Prediction of variant effect scores for missense SAVs & 2020 & Protein & 7,239 & \makecell{Human, Yeast,\\ Bacteria, Virus} & 46\\
    BEND \citep{marin2023bend} & Binary classification of non-coding SNPs as effect/no effect. (1) Gene expression (2) Disease & 2023 & DNA & \makecell{(1) 105,263\\(2) 295,495} & Human & 13\\
    Updated benchmarking of variant effect predictors using deep mutational scanning \citep{livesey2023updated} & Prediction of variant effect scores for missense SAVs & 2023 & Protein & 9,310 & Human & 55\\
    Genome Understanding Evaluation \citep{zhou2024dnabert-2:} & Classification of SARS-CoV-2 variant pathogenicity & 2024 & RNA & 91,669 & SARS-CoV-2 &\\
    Genomic Long-Range Benchmark \citep{kao2024advancing} & Prediction of SNP effect on gene expression & 2024 & DNAs & \makecell{(1) \citep{avsec2021effective}: 97,563.\\ (2) \citep{benegas2023gpn-msa}: 39,652.\\ (3) \citep{benegas2023gpn-msa}: 2,321,473.} & Human & 3\\
    Benchmarking computational variant effect predictors by their ability to infer human traits \citep{tabet2024benchmarking} & Prediction of functional scores for rare-disease-associated variants in the human genome & 2024 & DNA & 100,000 & Human & 24\\
\bottomrule
\end{tabular*}
\end{table*}

\begin{table*}
    \caption{Most common datasets used in papers on language modelling for variant effect prediction. ClinVar, a large open-access database of human genomic variants, is the most widely used. Data sourced from ClinVar has been employed for both training and evaluation. Pub. Year = Publication Year. No. Citations = Overall number of citations as per Google Scholar. Papers = Papers in this review using the dataset. * While the paper reporting the creation of the dataset \citep{avsec2021effective} has 835 citations, it was not possible to determine the number of citations for the dataset itself. \label{tab:datasets}}
    \begin{tabular*}{\linewidth}{@{\extracolsep{\fill}}p{0.2\linewidth}lp{0.2\linewidth}p{0.1\linewidth}llp{0.1\linewidth}l@{\extracolsep{\fill}}}
    \toprule
    Dataset & Data Type & Description & Size & Pub. Year & No. Citations & Papers & Open-Access\\
    \midrule
    ClinVar \citep{landrum2016clinvar} & DNA & ``...germline and somatic variants of any size, type or genomic location.'' \citep{landrum2016clinvar} & 500,000 variants \citep{landrum2018clinvar} & 2016 & 2,875 & \cite{brandes2023genome-wide, lin2024enhancing, fan2022shine:, liu2022protein, benegas2023gpn-msa, hidayat2023utilizing, wild2024dna, yang2022integrating, james2023deep, ji2021dnabert, gao2023epigept:, cheng2023accurate, zhan2024dyna:, danzi2023deep, lafita2024machine, zhou2022unsupervised, yan2024transefvp:, manfredi2022e-snpsgo:, dalla2024nucleotide, joshi2025augmented} & Yes\\
    gnomAD \citep{karczewski2020mutational} & DNA & Genome and exome sequences & 76,215 genomes, 730,947 exomes & 2020 & 8,243 & \cite{jiang2023deciphering, brandes2023genome-wide, lin2024enhancing, fan2022shine:, benegas2023gpn-msa, danzi2023deep} & Yes\\
    Human Gene Mutation Database (HGMD) \citep{stenson2020human} & DNA & ``...all known gene lesions underlying human inherited disease...'' \citep{stenson2020human} & 291329 entries (free version) 510804 entries (paid version) & 2020 & 1008 & \cite{jiang2023deciphering, brandes2023genome-wide, pejaver2020inferring, yang2022integrating, dalla2024nucleotide} & \makecell{Yes - Free version\\excluding past three\\years' data.}\\
    UniProt \citep{apweiler2004uniprot} & Protein & Protein sequences + annotations, including functional information & 253,206,171 entries & 2004 & 2,900 & \cite{shin2021protein, olenyi2022lambdapp:, manfredi2022e-snpsgo:, hidayat2023utilizing} & Yes\\
    CAGI5 Regulation Saturation \citep{shigaki2019integration} & DNA & Non-coding SNPs + effect scores & 175,000 variants across 9 promoters and 5 enhancers & 2019 & 56 & \cite{avsec2021effective, tang2023building} & Yes\\
    Variant Effect Causal eQTL \citep{avsec2021effective} & DNA & Non-coding SNPs + effect scores & 97,563 variants \citep{kao2024advancing} & 2021 & Unknown* & \cite{avsec2021effective, schiff2024caduceus:} & Yes\\
    \bottomrule
    \end{tabular*}
\end{table*}

\end{landscape}

\subsection{Beyond the Transformer}

\begin{figure}
    \centering
    \includegraphics[width=\linewidth]{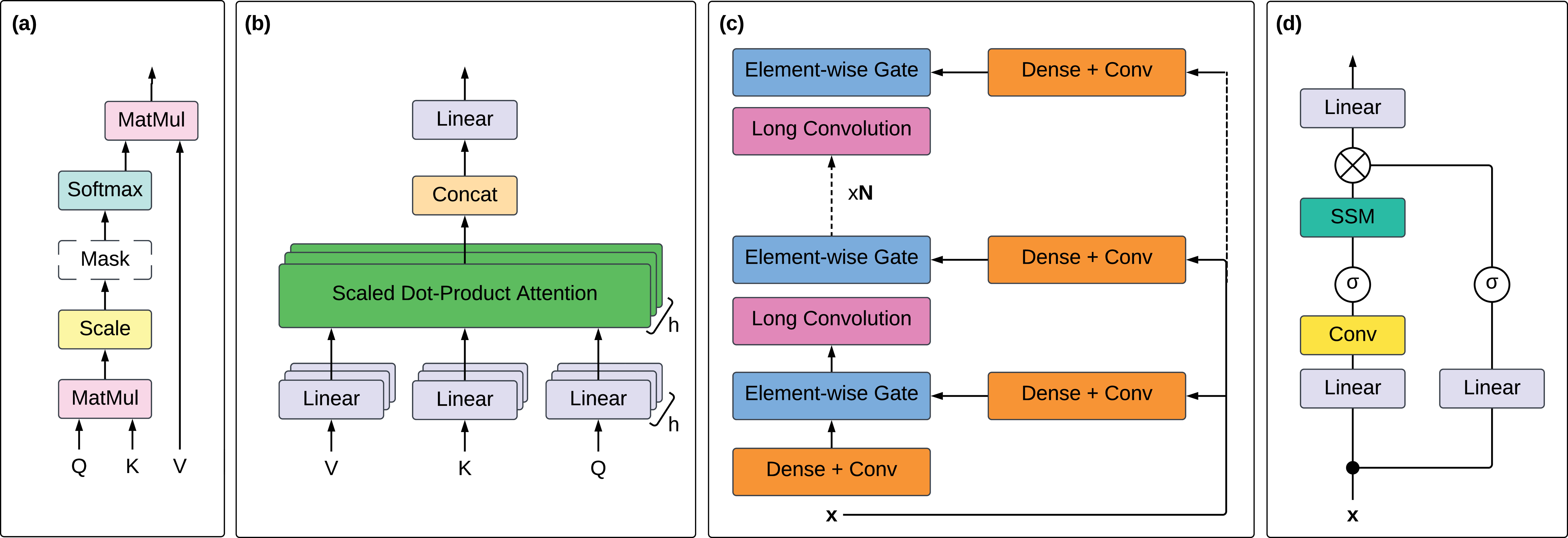}
    \caption{Comparison of the self-attention mechanism and alternatives. (a) Scaled dot-product attention, as shown in \citep{avsec2021effective}. The attention mechanism is applied simultaneously to a set of queries $Q$, with keys $K$ and values $V$. Hence, the output matrix is computed as: $Attention(Q, K, V) = softmax(QK^T/\sqrt(d_k))V$. \textit{MatMul} = matrix multiplication. The \textit{Mask} between the Scale and Softmax is used only in the decoder to preserve the auto-regressive property, by preventing the flow of data from right to left \citep{vaswani2017attentionisallyouneed}. (b) Multi-head attention, as shown in \citep{avsec2021effective}. The presence of $h$ heads indicates that $h$ attention layers run in parallel. (c) Hyena operator of order $N$, as shown in \citep{nguyen2023hyenadna}. Combinations of dense layers and convolutions are applied to the input; the resulting projections are then fed to the element-wise gate layers. An MLP is used to implicitly parameterise the long convolutions, hence producing the convolutional filters \citep{nguyen2023hyenadna}. \textbf{x} indicates the input. (d) Mamba operator, adapted from \citep{gu2024mambalineartimesequencemodeling}. The Mamba operator combines a state space model (SSM) with an MLP. \textbf{x} indicates the input. For the activation function $\sigma$, either a sigmoid linear unit \citep{hendrycks2016gaussian} or Swish \citep{ramachandran2017searching} is used.\\
    \textbf{Alt text:} Diagrams of attention mechanism and alternatives, with sub-figures labelled a to d.}
    \label{fig:attention_and_alternatives}
\end{figure}

In recent years, extensions and alternatives to the self-attention mechanism have been developed in order to tackle the high computational cost currently associated with training Transformer-based LLMs. The timeline of these emerging technologies is displayed in Figure \ref{timeline2}. Figure \ref{fig:attention_and_alternatives} provides a visual representation of the self-attention mechanism (a), multi-head self-attention (b), and the two major alternatives (c, d).
The first such approach to gain traction was the Hyena operator (Figure \ref{fig:attention_and_alternatives}(c)), which was developed in 2023 as a direct replacement for the self-attention mechanism. Using a recurrence of multiplicative gating interactions and long convolutions \citep{poli2023hyena}, this approach scales linearly with sequence length, unlike the attention mechanism, which scales quadratically. Thus, the Hyena operator is 100 times faster than attention at a sequence length of 100,000 bases, while delivering similar results \citep{nguyen2023hyenadna}. This operator forms the basis of HyenaDNA \citep{nguyen2023hyenadna}, a foundation model for DNA, which has achieved excellent results on tasks such as chromatin profile prediction and species classification.
An alternative replacement for the attention mechanism is the state space model-based Mamba operator \citep{gu2024mambalineartimesequencemodeling} (Figure \ref{fig:attention_and_alternatives}(d)). Unlike conventional state space models, which experience performance bottlenecks due to repeated matrix multiplications, Mamba uses a structured state space sequence (S4) model, which overcomes this by employing matrix diagonalisation. Additionally, the Mamba-based model outperformed HyenaDNA on a species classification task while using the same number of parameters, suggesting that Mamba models biological sequences more accurately and efficiently.
Despite several developments in post-Transformer methods, few of these models have been applied to variant effect prediction (Table \ref{tab:post_transformer_models}).

One of the first post-Transformer models applied to variant effect prediction is Caduceus \citep{schiff2024caduceus:}, which is based on the Mamba operator \citep{gu2024mambalineartimesequencemodeling}. The implementation leverages the reverse complement (RC) nature of the two strands in a double-helix DNA structure, recognising that both strands contain semantically equivalent information. The Mamba operator is applied twice, once to the original DNA sequence, and again to a reversed copy of the sequence; the parameters are shared between these two applications to increase efficiency. This double application of the operator is termed BiMamba, and is used as the basis of the MambaDNA block, which additionally defines an RC mathematical operation to re-combine the forward and reverse sequences.
The performance of the model was evaluated on a non-coding variant effect prediction dataset \citep{avsec2021effective}, and was compared with the state-of-the-art foundation models HyenaDNA \citep{nguyen2023hyenadna} and Nucleotide Transformer \citep{dalla2024nucleotide}. Caduceus outperformed both state-of-the-art models, achieving an AUROC of 0.68 on variants that were 0-30 kbp (kilo-base-pairs) from the nearest transcription start site (TSS). However, performance degraded with increasing distance of the variant from the nearest TSS, with the AUROC decreasing to 0.61 for variants at a distance of 100+ kbp. Notably, Caduceus was able to surpass the performance of Nucleotide Transformer v2 using only a fraction of the parameters (7.7M compared to the Nucleotide Transformer's 500M).

The other notable example of a post-Transformer model applied to variant effect prediction is Evo \citep{nguyen2024evo}. This is a hybrid Transformer-Hyena model, where Hyena operators are combined with multi-head self-attention to improve performance on long sequences; this approach is termed StripedHyena \citep{poli2023striped}. Evo was pre-trained on a prokaryotic whole-genome dataset of 300 billion nucleotides, resulting in a model with 7 billion parameters, which can handle a context length of up to 131,072 nucleotides \citep{nguyen2024evo}. Analysis during training showed that the model scaled far better with sequence length compared to state-of-the-art Transformer models; while the Transformer-based models scaled quadratically with sequence length, the scaling of Evo was almost linear. However, the training was highly resource-intensive, with the first stage taking two weeks across 64 GPUs, and the second stage taking a further two weeks across 128 GPUs. Hence, the availability of the pre-trained model is a major contribution of this work, as it can be applied to different tasks without requiring re-training from scratch.
Evo’s performance on variant effect prediction was tested across two tasks. Firstly, the prediction of variant effects on bacterial protein fitness. The Spearman correlation between the experimental and predicted fitness values was 0.45, underperforming compared to state-of-the-art models, including Nucleotide Transformer \citep{dalla2024nucleotide} and RNA-FM \citep{chen2022interpretable}, which achieved correlation values between 0.5 and 0.55 \citep{nguyen2024evo}. The second task was the prediction of variant effects on non-coding RNA fitness, in which Evo achieved a Spearman's correlation of 0.27 between its predictions and the true values. While this exceeds state-of-the-art models, which achieved a correlation of less than 0.2 on the same task, the performance indicates that further research is required to produce a model that can accurately predict variant fitness in non-coding RNA. Evo was also tested on predicting mutational effects on human protein fitness, however, these experiments were unsuccessful; it was hypothesised that this may be due to the model being trained only on prokaryotic sequences, without any human samples.

These models have achieved mixed results. While in some cases, they have matched or exceeded state-of-the-art performance while reducing the number of model parameters required, the state-of-the-art models demonstrate limited ability to predict variant effects. While improvements in computational efficiency have been achieved using models such as Caduceus, this remains an area requiring further attention. For instance, Evo has achieved results exceeding the current state-of-the-art, and the pre-trained model has been made available, however it would be necessary to undertake the resource-intensive pre-training stage again in order to make it suitable for use on the human genome. These outcomes indicate that significant further research is required to ascertain whether these technologies are indeed effective for modelling genetic sequences.

\begin{table*}[t]
\caption{Summary of post-Transformer large language models for variant effect prediction. See Table for code/data availability.\label{tab:post_transformer_models}}
\tabcolsep=2pt
\begin{tabular*}{\textwidth}{@{\extracolsep{\fill}}lp{0.25\textwidth}lp{0.25\textwidth}ll@{\extracolsep{\fill}}}
\toprule
Paper & Task & Year & Architecture & Data Type & Variant Type\\
\midrule
    \citep{schiff2024caduceus:}* & \makecell{Non-coding variant effect\\prediction} & 2024 & Caduceus; based on Mamba\newline\citep{gu2024mambalineartimesequencemodeling} & DNA & Non-coding\\
    \citep{nguyen2024evo} & \makecell{(1) Predicting mutational effects\\on bacterial protein fitness\\(2) Predicting mutational effects\\on non-coding RNA fitness.} & 2024 & \makecell{Evo, based on StripedHyena\\\citep{poli2023striped}} & DNA, RNA, Protein & \makecell{Coding,\\Non-coding}\\
\bottomrule
\end{tabular*}
\end{table*}

\section{Model Evaluation}

This section details the approaches to model evaluation for language models in variant effect prediction. First, the main datasets used in the field are reviewed. Then, benchmarking studies are evaluated. Finally, relevant metrics and evaluation protocols are surveyed.

\subsection{Datasets \& Benchmarking}

A considerable challenge in the field is the difficulty of accurately comparing different models. The papers reviewed employ a variety of datasets and metrics, which seldom align. Even in the case of datasets or tasks that are used to assess multiple models, different papers select different subsets of the dataset, or apply different metrics to measure model performance. This makes it challenging to compare the performance of various methods, and hence can obscure the effect of different architectures on prediction quality. Hence, there is a pressing need for benchmarks that can enable comparison of models.

\begin{figure}
    \centering
    \includegraphics[width=\linewidth]{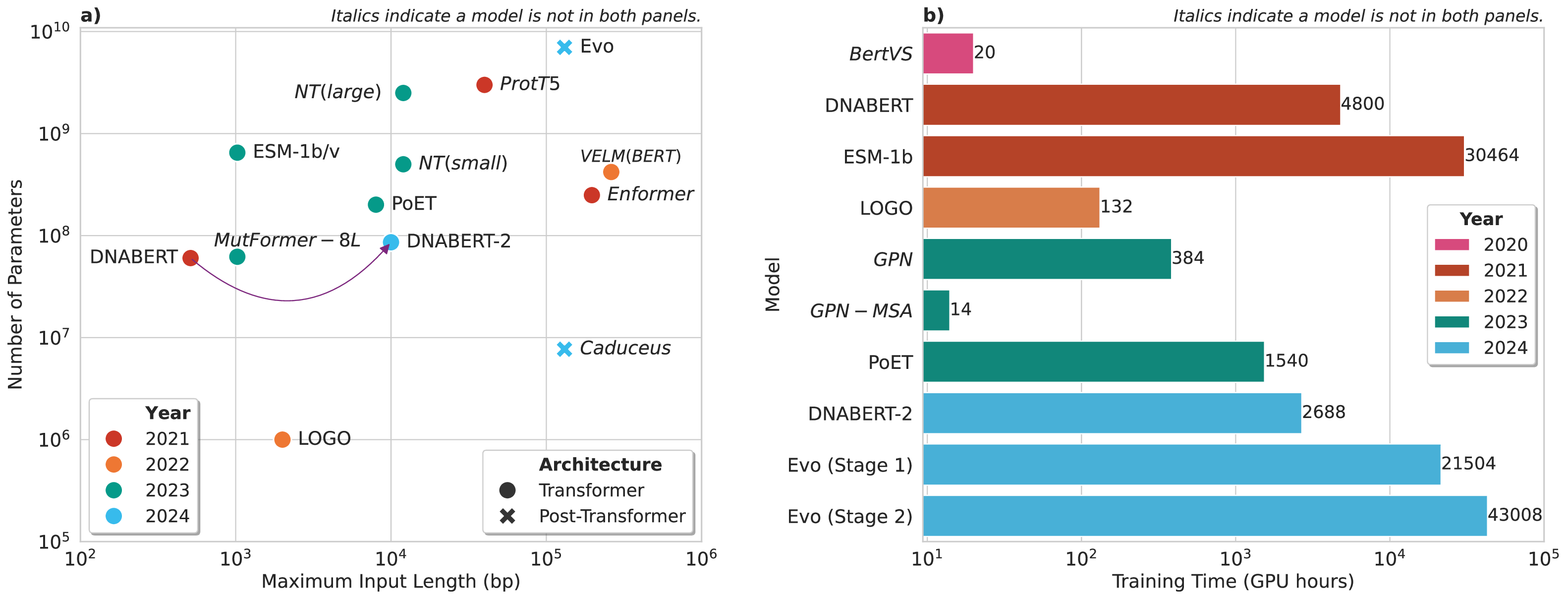}
    \caption{Input sequence length, number of parameters, and training time for models which have reported these statistics in the original papers. \textbf{(a)} Maximum input sequence length (x-axis) and number of parameters (y-axis) as reported in original papers for each model. The model names are indicated on the chart.  There is no clear trend shown over time. Compared to the majority of Transformer-based models, Caduceus, a Mamba-based model, has far fewer parameters and can handle longer input sequences. \textbf{(b)} Training time in GPU hours for state-of-the-art LLMs. GPU hours = number of hours x number of GPUs. In general, the training time required for LLMs has increased over the years. However, DNABERT and ESM-1b are outliers, having very high training times; this is likely due to the fact that both are foundation models, which were trained on very large datasets. GPN-MSA is another outlier, and has a particularly low training time, likely due to the use of retrieval augmented processing \citep{borgeaud2022improving} to increase computational efficiency \citep{benegas2023gpn-msa}.\\
    \textbf{Alt text:} Graphs showing the relationship between maximum input length, number of model parameters, training time, and year of publication, with sub-figures labelled a and b.\label{fig:len_params_time}}
\end{figure}

Table \ref{tab:datasets} summarises the main datasets used in the papers reviewed above; datasets used across multiple papers were identified, and their characteristics summarised.
The most impactful DNA database in the field is ClinVar \citep{landrum2016clinvar}; its coverage of many different variants across the whole human genome makes it suitable for training and evaluating a wide range of models. Two other similar databases are also very popular; gnomAD \citep{karczewski2020mutational} and the Human Gene Mutation Database \citep{stenson2020human}. The former is unique due to its inclusion of several different ancestry groups from around the globe. Additionally, an equivalent for proteins exists in the form of UniProt, which contains over 200,000,000 protein sequences and annotations, and is exploited by several protein language models. Though these databases are used across many scientific articles, it is rare for different models to be evaluated on the same subset of a database. As shown in Tables \ref{supp:neural}, \ref{supp:transformer_models}, and \ref{supp:post_transformer_models}, the datasets used in the field are numerous, vary significantly across papers, and frequently are not open-access. Even the datasets that have been the most popular, such as the CAGI5 Regulation Saturation \citep{shigaki2019integration} and Variant Effect Causal eQTL \citep{avsec2021effective} datasets, have only been employed across a small number of papers (Table \ref{tab:datasets}).
These constraints make it challenging to compare the performance of different models, as their datasets may vary significantly in the type of data or category of task. Addressing this limitation either requires the community to agree on a set of datasets on which to evaluate new models, or the compilation of a framework or dataset that covers several different tasks. A source of inspiration should be the Critical Assessment of Structure Prediction (CASP) \citep{shortle1995protein}, a recurring set of experiments to determine the state-of-the-art in protein structure prediction methods. Every two years since 1994, research groups worldwide have been encouraged to submit results to ensure that a thorough and complete review of existing methods is conducted. The experiments provide a method for researchers across the community to evaluate their models on a common dataset, and provides several categories of tasks on which models can be assessed. This format could be highly applicable for the variant effect prediction community. A regular competition or community experiment comprising multiple categories of variant effect prediction tasks on varied context lengths would be invaluable in determining the state-of-the-art and deciding the course of future research. Furthermore, input from the clinical community on desired standards and ideal tasks could be used to assess the real-world applicability of such models.

Currently, benchmarking studies in the field are limited. However, significant progress has been made by Livesey and Marsh at the University of Edinburgh in benchmarking protein variant effect predictors, with two successive studies published in 2020 \citep{livesey2020using} and 2023 \citep{livesey2023updated}. They provide a comprehensive review of protein variant effect predictors at the time of publication, comparing their performance on deep mutational scanning datasets of human proteins, and ranking the models based on their results. The difference between the two articles highlights the progress in protein language modelling over the early 2020s. While the 2020 study identified DeepSequence \citep{riesselman2018deep}, a non-language-modelling, probabilistic model as the best variant effect predictor for proteins, the 2023 one revealed that LLM methods such as ESM-1v \citep{meierlanguage} produced even better results. Another notable finding was the increase in data availability; in the 2023 study, there were over twice as many datasets available on which to evaluate the models. A particular strength of this study is that models were compared across multiple metrics - AUROC, AUBPRC, and correlation; the benefits of this are discussed further in the metrics section. Overall, these two studies provide a thorough review of the existing models for protein variant effect prediction. However, language modelling specific aspects are not explored, as deep learning models of various methodologies are assessed.

Though variant-specific benchmarks are scarce, variant effect prediction tasks are included in some benchmarking studies that evaluate the performance of LLMs on genomic modelling in general. For instance, the Genome Understanding Evaluation benchmark \citep{zhou2024dnabert-2:} consists of genomic modelling tasks across multiple species, including the classification of SARS-Cov-2 variants based on sequences of 1000 base pairs (bp) in length. Comparison of DNABERT-2 with several different versions of DNABERT and Nucleotide Transformer showed that a version of the Nucleotide Transformer pre-trained on multispecies data performed best, while DNABERT-2 was close behind (accuracies of 73.04\% and 71.21\% respectively). A complementary study is the Genomics Long-Range Benchmark \citep{kao2024advancing}, which evaluates model performance specifically on genomics tasks requiring modelling of long-range dependencies, and includes the prediction of SNP effect on gene expression, using data derived from \citep{avsec2021effective}. It was discovered that increasing context length improved models' ability for variant effect prediction. Additionally, models with longer context lengths were able to more accurately predict the effects of variants further from the nearest transcription start site (TSS). Indeed, Enformer outperformed more recent models such as Nucleotide Transformer and HyenaDNA due to its ability to handle longer context.

While past benchmarks focused on the quality of predictions, there is also a need to understand and compare the computational cost of variant effect prediction models. Recent research has highlighted the immense impact of deep learning technologies on the natural environment, from carbon emissions to water consumption \citep{huang2024white, jiang2024preventing}. Transformer-based LLMs are a significant culprit due to the quadratic scaling of the attention mechanism with context length. The computational cost of training on large datasets can be extensive; as shown in Figure \ref{fig:len_params_time}, training can span across days or weeks, using multiple GPUs. Though large foundation models such as DNABERT and ESM-1b are particularly computationally expensive to train, the training time in general has increased since 2020. However, training is not the only computational expense associated with LLMs; while training only occurs once, inference occurs repeatedly, with the frequency depending on the application of the LLM. For instance, ChatGPT was visited over three billion times in December 2024 \citep{similarweb2024chatgpt}. Hence, since the total inference cost over time can match or exceed the training cost, it is crucial to understand and reduce its impact in the pursuit of environmentally conscious models.
Table \ref{tab:inference_time} lists the inference time as per the original paper for each model. Notably, not all LLM methods have high inference time, and many improve on traditional methods. Additionally, recent methods have aimed to perform inference on consumer-grade machines rather than high-specification GPUs, hence making the models more accessible to run in clinical settings. For instance, VespaG \citep{marquet2024vespag:} took only 5.7 seconds on a 12-core CPU to make predictions for 73 unique proteins from ProteinGym \citep{notin2023proteingym}, while a non-LLM method, GEMME \citep{laine2019gemme}, took 1.27 hours to perform the same task on the same hardware.
However, inference time is still far less frequently reported than training time - the only models for which this is reported are listed in Table \ref{tab:inference_time}. Hence, it is also challenging to compare existing methods based on this criterion.

\begin{table*}[t]
\caption{Comparison of reported inference time for LLM methods.\label{tab:inference_time}}
\tabcolsep=0pt
\begin{tabular*}{\textwidth}{@{\extracolsep{\fill}}lllll@{\extracolsep{\fill}}}
\toprule%
Model & Publication Year & Transformer-based & CPU/GPU & Inference Time\\
\midrule
E-SNP\&GO \citep{manfredi2022e-snpsgo:} & 2022 & Yes & 1 x 12-core CPU & 12.464 seconds per variant \\
\makecell{VariPred \citep{lin2024enhancing}\\(based on ESM-2 \citep{lin2023evolutionary})} & 2024 & Yes & 1 x GPU - 12GB Nvidia GTX 1010Ti & 0.360 seconds per variant \\
VespaG \citep{marquet2024vespag:} & 2024 & Yes & 1 x 12-core CPU & 0.078 seconds per protein \\
\bottomrule
\end{tabular*}
\end{table*}

\subsection{Metrics}
Three main categories of metrics are used to evaluate computational variant effect predictors.
The first such category contains metrics that align with those used for standard machine learning models, and use true and false positive rates to evaluate the predictions. These include area under the receiver operator characteristic curve (AUROC) \citep{sun2023structure-informed, benegas2023gpn-msa}, accuracy \citep{jiang2023deciphering, hidayat2023utilizing}, precision \citep{dampier2022hiv}, recall \citep{dampier2022hiv}, and F1-score \citep{yang2022integrating, zhou2024dnabert-2:}.

The second category of metrics assesses the relationship between the true values and those predicted by the model. In cases where a numerical value such as variant effect score is predicted, this is done by calculating the correlation between the two. Spearman’s rank correlation coefficient is most frequently used \citep{meierlanguage, derbel2023accurate, hidayat2023utilizing, rives2021biological}; however, some papers also use Pearson’s \citep{dunham2023high-throughput, avsec2021effective} correlation coefficient. All such metrics used in the reviewed papers are summarised in Table \ref{tab:metrics}. While all of these metrics measure the agreement between the true and predicted values, they each measure this in a different way. For instance, Pearson's correlation coefficient assesses whether there is a linear relationship between the two, while Spearman's correlation coefficient determines whether a monotonic relationship exists.
A unique case is Matthews’ correlation coefficient (MCC) \citep{matthews1975comparison}, which is used to evaluate the agreement between the true and predicted classes in a classification problem \citep{zhou2024dnabert-2:, lin2024enhancing}. Unlike accuracy or AUROC, it takes into account all four aspects of a confusion matrix (true and false positive rates, and true and false negative rates), hence better representing the overall quality of predictions produced by the model \citep{chicco2023matthews}.

\begin{table*}[t]
\caption{Metrics used for assessing the relationship between the values predicted by the model and the true values. Pearson's, Spearman's and Jaccard metrics are used for prediction of numerical values. Matthews' is used for classification.\label{tab:metrics}}
\tabcolsep=2pt
\begin{tabular*}{\textwidth}{@{\extracolsep{\fill}}llll@{\extracolsep{\fill}}}
\toprule
Metric & Measures & Range & Note\\
\midrule
Pearson's correlation coefficient \citep{pearson1903mathematical} & Linear relationship & (Strong negative) -1 to 1 (Strong positive) & 0 = No relationship\\
Spearman's correlation coefficient \citep{spearman1961proof} & Monotonic relationship & (Strong negative) -1 to 1 (Strong positive) & 0 = No relationship\\
Matthews' correlation coefficient \citep{matthews1975comparison} & Agreement of classes & (All inverse) -1 to 1 (All correct) & 0 = No agreement\\
Jaccard similarity index \citep{jaccard1901etude} & Similarity & (No similarity) 0 to 1 (Complete similarity) &\\
\bottomrule
\end{tabular*}
\end{table*}

To compare the agreement of these two categories of metrics, a simple meta-predictor was created by using the pre-trained Enformer model \citep{avsec2021effective} to generate embeddings from SNPs in the ncVarDB \citep{biggs2020ncvardb}, and using a simple machine learning classifier on top to perform a binary pathogenicity classification. The results of the different models tested are displayed in Table \ref{tab:metrics_comparison}. It must be noted that, while Random Forest and Gradient Boosting achieved the same accuracy, their AUROC and MCC were different. Additionally, the MCC achieved using SVM with a linear kernel is very similar to that achieved using Random Forest, despite the latter having higher accuracy and AUROC values. These results demonstrate the importance of evaluating and comparing models across these different dimensions in order to fully understand the differences and determine the state-of-the-art.

\begin{table}[t]
\caption{Comparison of metrics for models performing variant pathogenicity classification on SNPs from ncVarDB \citep{biggs2020ncvardb}, using embeddings extracted from Enformer \citep{avsec2021effective}. MCC = Matthews' correlation coefficient. \label{tab:metrics_comparison}}
\tabcolsep=2pt
\begin{tabular*}{\columnwidth}{@{\extracolsep{\fill}}llll@{\extracolsep{\fill}}}
\toprule
Model & Accuracy & AUROC & MCC\\
\midrule
SVM (RBF kernel) & 69.4\% & 0.725 & 0.516\\
SVM (linear kernel) & 73.6\% & 0.763 & 0.574\\
Random Forest & 77.5\% & 0.781 & 0.572\\
Gradient Boosting & 77.5\% & 0.778 & 0.558\\
\bottomrule
\end{tabular*}
\end{table}

Further to the more generic metrics in the first two categories, a significant and specific metric in NLP is \textit{perplexity} \citep{jelinek1977perplexity}. Indeed, language models represent sequences by calculating the probability of each token based on the context from previous tokens. The perplexity is calculated by taking the inverse probability assigned to each token in a given set of data and normalising it by the number of words as shown in Equation \ref{1eq:ppl} for a dataset $W = w_{1}w_{2}...w_{N}$ \citep{jurafskymartin2024speech}. 
\begin{equation}
    perplexity(W) = P(w_{1}w_{2}...w_{N})^{-{1/N}} \label{1eq:ppl}   
\end{equation}

For a given model, a lower perplexity indicates an enhanced ability to predict the next token of a sequence. Perplexity can be calculated continuously throughout during the pretraining stage to identify the optimal number of parameters \citep{nguyen2024evo}. However, while an improvement in perplexity often correlates with an improvement in performance on downstream tasks, this relationship is not guaranteed, and hence, further evaluation metrics are required to directly evaluate the performance of the model on the task of interest \citep{jurafskymartin2024speech,meister2021lmevaluation}. For instance, though Evo achieved a lower pretraining perplexity compared to Transformer-based models, the latter still achieved better Spearman's correlation between true and predicted values when predicting bacterial protein fitness \citep{nguyen2024evo}.

Beyond perplexity, no further NLP-specific metrics have been used to evaluate variant effect predictors based on language models. However, many such metrics have been developed to evaluate the ability to model natural languages, such as ROUGE \citep{lin2004rouge} and its variants, and a variety of semantic embedding-based metrics \citep{rus2012optimal, forgues2014bootstrapping}. Moreover, recent papers have investigated the use of semantic similarity for assessing the ability of LLMs to appropriately model natural languages. Of particular interest is a 2024 paper in which the ability of an encoder to model substitution of a word with a synonym or antonym is tested \citep{xu2024reasoning}; this concept could be extended to genetic language modelling, and hence evaluate the ability of an encoder to model substitution of a nucleotide. Despite the ability of non-NLP-specific metrics to evaluate the results of a model, they have no ability to assess the quality of language modelling or understand the underlying logic. Hence, to fully understand LLM performance, standard metrics must be combined with NLP-specific metrics.

While there are several metrics to assess the quality of model predictions, looking solely at the values of these metrics does not take into account other key aspects of a model, including computational cost. Though modifications such as including additional features in the training data, or increasing the size of the model, can enhance the predictive performance, they can also lead to a significantly higher computational cost. This calls into question the extent to which an increase in computational cost is justified for a corresponding increase in prediction quality \citep{thompson2020computational}. For instance, usage of Pareto optimality has been adopted to attempt to select models with an appropriate trade-off between accuracy and inference latency \citep{justus2018predicting}. In future, it would be very valuable to define a metric to combine the information from each of the three categories above with data regarding computational cost.

\section{Discussion}
The advent of the Transformer model in 2017 led to a paradigm shift in NLP and its applications to various fields, including the prediction of biological variant effects. Transformer-based language models have achieved mixed results in this area; while some models excel, others fail to make accurate predictions.
Another significant limitation of Transformers is the overwhelming computational cost associated with training and inference due to the quadratic scaling of the cost of the attention mechanism with sequence length. Research to address this has led to the development of several attention alternatives such as Mamba and Hyena. While these have garnered much attention in the LLM field, their capacity for variant effect prediction has not yet been fully explored, with only two models being used for this application so far. Additionally, Transformer-based models are still being proposed for variant effect prediction, as recently as early 2025 \citep{joshi2025augmented}, demonstrating that this technology remains competitive.

The models produced to date have focused largely on single-nucleotide substitutions within proteins, or protein-coding regions of the human genome, often achieving promising results. However, there has been very little work on multiple base-pair variants, or non-substitution variant types, such as indels. Furthermore, while there has been extensive work on modelling DNA and protein sequences, there has been limited work on human RNA, despite the known associations between RNA variants and disease \citep{pickrell2010understanding, manning2017roles}. Moreover, though extensive research has been conducted on the effects of variation within the human genome, very few recent studies have investigated the effects of variants in pathogenic organisms and viruses with a high disease burden. In particular, only two studies \citep{zhou2024dnabert-2:, gao2023epigept:} have looked at the mutational effects of SARS-CoV-2, which had a devastating impact on human health during the Covid-19 pandemic. Still, some work has been conducted on using deep learning to viral mutation data to predict individual risk \citep{tai2022machine} and the possibility of drug resistance \citep{das2022effective}. Moreover, given that LLMs have already demonstrated effectiveness in modeling HIV \citep{dampier2022hiv}, they could potentially enhance results in this area.

Despite significant advancements in recent years, the field still faces several limitations. Many of the most prominent challenges are related to data rather than model architectures.
A common issue observed among computational variant effect predictors is \textit{type 2 data circularity}. Studies found that, in many cases, all variants within a particular gene are recorded with the same label (pathogenic or benign) across multiple different variant databases. This leads to models trained on this data performing well on known variants in known genes, but poorly on de novo variants for newly identified risk genes \citep{grimm2015evaluation}. Though a benchmarking study investigating this issue found that traditional machine learning models were the most prone to suffering from this issue, only one LLM (an ESM variant) was tested, hence it is possible that others may still be at risk of suffering from this issue \citep{lin2024enhancing}. Therefore, it may be of interest to include such a test in future LLM benchmarking studies.

Another significant data-related challenge is that of demographic bias. Many large genomics datasets, such as UK Biobank, contain data largely from individuals of White European descent \citep{biobankancestry}. This poses a concern, as several mutations related to Mendelian diseases, including sickle-cell anemia and Tay-Sachs disease, have been shown to differ significantly in prominence across different groups \citep{lu2014personalized, prohaska2019human}. Hence, training on an ancestrally homogenous dataset risks the loss of valuable features when modelling the human genome, and can lead to poor generalisation of models across different ancestral groups. The computational healthcare field has largely continued to uphold existing biases against underserved groups, with some widely-used algorithms displaying clear racial bias \citep{obermeyer2019dissecting}. As the field moves forward into an era where  algorithms play an increasingly pivotal role in shaping personalised medicine, it is crucial to prioritise equity in future developments to ensure fair and unbiased outcomes for all.

In addition to addressing dataset composition, the privacy of patient data is another key consideration when using LLMs for healthcare-related applications. As LLMs have already demonstrated their ability to identify sensitive information in documents such as electronic health records \citep{chen2019generation, liu2023deid}, this raises concerns around accidental patient identification via training data. Genomic data must be treated as particularly sensitive, due to the possibility of identifying not only an individual, but also their familial relationships, and links to specific traits or diseases \citep{naveed2015privacy}. This is of particular concern in rare disease research, where access to data on diseases experienced by only a handful of individuals increases the risk of individuals being identified. Although privacy solutions for genomic data sharing are being rapidly explored and developed \citep{grishin2019data, bonomi2020privacy}, it is crucial to consider these through the lens of LLMs and the handling of data by those who develop these models.
Indeed, LLMs can be susceptible to Membership Inference Attacks (MIA) \citep{shokri2017membership} and User Inference Attacks (UIA) \citep{kandpal2023user}. MIA aims to determine whether a given data record is present in the training data of an LLM, and is conducted by creating an adversarial model to recognise the differences in an LLM's response to its training data and its response to other samples. Recent research has shown that such attacks are effective on clinical language models, with samples from individuals with rare diseases being at greater risk of privacy leakage \citep{jagannatha2021membership}. On the other hand, UIA attempts to ascertain whether an individual's data was used in fine-tuning an LLM. While MIA threatens the privacy of individual samples, UIA puts the privacy of users who have contributed multiple samples at risk \citep{kandpal2023user}. Both sets of attacks can severely compromise patient data privacy, and can lead to the revelation of sensitive information about participants. However, tests on MIA and UIA have not yet been applied to genomic language models, and the latter has not yet been tested for any clinical LLMs. Hence, a framework must be created for testing the resiliency of state-of-the-art models in the field against such attacks. Crucially, these tests must be performed before models are adopted into clinical settings, to avoid putting patients at risk.

\subsection{Future Trends}
Due to the significant training and inference costs associated with Transformer-based LLMs, many recent studies have focused on creating more computationally efficient models, either using Transformers, or substituting the attention mechanism with alternative operators such as Hyena or Mamba. Although the advent of small language models (SLMs)\citep{zhang2024tinyllama} has advanced this area of research in natural-language-based LLMs, they have not yet been applied to genetic sequences. A notable SLM is TinyLlama \citep{zhang2024tinyllama}, which utilises the same architecture and tokeniser as Llama2 \citep{touvron2023llama}, while leveraging novel computational methods such as FlashAttention \citep{dao2023flashattention} to create a model with fewer parameters and increased computational efficiency compared to state-of-the-art LLMs. SLMs have already demonstrated impressive performance in text classification \citep{dehan2025tinyllm} and text-based health monitoring \citep{wang2024efficient}, matching or exceeding the results achieved using LLMs. These findings underscore the potential of SLMs in future research, and suggest that they may be an interesting avenue of advancement for biological language modelling also.

Though development of SLMs is on the horizon, LLMs continue to be widely used. Recent papers have shown a trend towards the use of foundation models, which are pre-trained on a large corpus of data and can be fine-tuned for a wide range of downstream tasks. For instance, eight separate papers in Table \ref{tab:transformer_models} base their models on the ESM-1b \citep{rives2021biological} foundation model. As the field aims to reduce computational cost, it is likely that foundation models will be even more widely used as an alternative to ab initio pre-training of new LLMs.

As the number of models in the field rapidly increases (\ref{fig:num_models}), often trained and evaluated on different datasets, it is becoming increasingly challenging to identify the true state-of-the-art. To address this rising need, the development of benchmarking datasets has accelerated since 2023, resulting in the creation of benchmarks such as the Genome Understanding Evaluation \citep{zhou2024dnabert-2:}. As interest in computational efficiency and model fairness grows, it is likely that future benchmarks will include methods to assess these features of models, and that such measures will become more significant when comparing models.
Moreover, though models may perform well during technical evaluations, is it crucial to define and adhere to specific standards in order to to discern their efficacy in clinical settings. For instance, in 2018, NHS England and the UK National Institute for Health and Care Excellence (NICE) developed an evidence standards framework \citep{nice2018esf} to provide guidance on the development and usage of digital health and care technologies. While this framework places a high emphasis on demonstrating valuable results and significant benefits to the target population, it is not specific to AI or LLM-based technologies, and hence does not detail any expectations for numerical results or other aspects of models.  It is therefore of the utmost importance that those in the computational field work closely with clinicians to decide appropriate standards for the performance of variant effect predictors, and implement strategies to bridge the gap between research and practice. Existing frameworks for models predicting individual prognosis or diagnosis include TRIPOD \citep{collins2015transparent}, which explores transparent reporting, and PROBAST \citep{wolff2019probast}, which estimates the risk of bias - these could be used to inform the creation of similar frameworks for language model -based variant effect predictors.

Alongside appropriate performance, the adoption of computational models in the clinical field requires the exploration of clinically relevant problems. While the bulk of work in the field has focused on the coding regions of the genome, research continues to uncover associations between non-coding variants and rare but highly impactful diseases in humans \citep{christensen2019rare, jiang2021variants, pagnamenta2023structural}. Thus, although there has recently been increasing interest in predicting the impact of human genetic variation in the non-coding regions, further computational exploration of the non-coding genome is required.
Furthermore, though current research focuses mainly on SNPs, diseases such as haemophilia have been linked to multiple base-pair variants or combinations of co-occurring SNPs \citep{bowen2002haemophilia, shetty2011challenges}. Very few papers exist on computational prediction of the effects of such variants \citep{liu2015predicting, holcomb2021new}, hence this is an area of great interest for future work.

\section{Conclusion}
Though language models have proven effective in modelling DNA, RNA, and protein sequences, their results on variant effect prediction tasks remain mixed. The best performance on these tasks has been achieved by large Transformer-based foundation models, pre-trained on large corpora of sequence data. However, such models incur a high computational cost in terms of training and inference. While this has begun to be addressed via the creation of alternatives and extensions to the attention mechanism, these have had limited use in bioinformatics thus far. Initial studies show that models based on these technologies, such as Caduceus and Evo, achieve results comparable to Transformer-based models while consuming less time and fewer resources for training and inference. Nevertheless, the state-of-the-art results for some tasks of importance, including non-coding variant effect prediction, require improvement.
Despite the substantial progress in the field in recent years, there are still a number of limitations that persist, including demographic bias in training datasets, and the limited work on variants spanning multiple base-pairs or situated in the non-coding regions of the genome.

\section{Competing interests}
No competing interest is declared.

\section{Author contributions statement}
M.H, J.N. and F.R. conceived the experiment(s),  M.H. conducted the review, analysed the results. M.H wrote initial draft, M.H. J.N. and F.R  reviewed and finalised the manuscript.

\section{Acknowledgments}
The authors thank the anonymous reviewers for their valuable suggestions. This work is supported by funds from the Kingston University Graduate Research School PhD Studentship.

\bibliographystyle{unsrt}
\bibliography{reference}

\section{Supplementary Information}

\begin{landscape}

\begin{table*}[ht!]
\caption{Code and data availability for neural language models in Table \ref{tab:neural}. \label{supp:neural}}
\tabcolsep=5pt%
\begin{tabular*}{\linewidth}{@{\extracolsep{\fill}}llp{0.35\textwidth}p{0.35\textwidth}p{0.15\textwidth}@{\extracolsep{\fill}}}
\toprule%
Paper & Data Source & Data Availability & Code Availability & Model\\
\midrule
\citep{kim2018mut2vec} & \makecell{\citep{hudson2010international},\\\citep{gonzalez2013intogen}} & N/A & N/A & N\\ 
\citep{pejaver2020inferring} & \makecell{\citep{stenson2020human},\\\citep{sherry2001dbsnp},\\\citep{mottaz2010easy}} & \url{http://mutpred.mutdb.org/wo_exclusive_hgmd_mp2_training_data.txt} & \url{https://github.com/vpejaver/mutpred2} & N \\
\citep{shin2021protein} & \makecell{\citep{riesselman2018deep},\\\citep{apweiler2004uniprot}} & \url{https://zenodo.org/records/4606785} & \url{https://github.com/debbiemarkslab/SeqDesign} & N \\
\citep{dunham2023high-throughput} & \makecell{\citep{alquraishi2019proteinnet},\\\citep{kryshtafovych2019critical}} & \url{https://zenodo.org/records/7621269} & N/A & \url{https://www.ebi.ac.uk/biostudies/studies/S-BSST732}\\
\citep{benegasdna} & \makecell{\citep{ncbi},\\\citep{1001genomes}} & N/A & \url{https://github.com/songlab-cal/gpn} & \url{https://huggingface.co/collections/songlab/gpn-653191edcb0270ed05ad2c3e}\\
\citep{tan2023multimodal} & \makecell{\citep{zhou2015predicting},\\\citep{chen2022sequence},\\\citep{biggs2020ncvardb}} & \url{https://zenodo.org/record/7975777} & \url{https://github.com/Shen-Lab/ncVarPred-1D3D} & N \\
\citep{cheng2023self-supervised} & \makecell{\citep{zhou2015predicting},\\\citep{avsec2021effective}} & N/A &  \url{https://github.com/wiedersehne/cdilDNA} & N \\
\bottomrule
\end{tabular*}
\end{table*}

\begin{longtable}{llp{0.3\linewidth}p{0.4\linewidth}l}
\tablecaption{Summary of Transformer-based language models for variant effect prediction. * = preprint.} \label{supp:transformer_models}

\\\hline \multicolumn{1}{c}{Paper} & \multicolumn{1}{c}{Data Source} & \multicolumn{1}{c}{Data Availability} & \multicolumn{1}{c}{Code Availability} & \multicolumn{1}{c}{Model} \\ \hline 
\endfirsthead

\multicolumn{5}{c}%
{{\bfseries \tablename\ \thetable{} -- continued from previous page}} \\
\hline \multicolumn{1}{c}{Paper} & \multicolumn{1}{c}{Data Source} & \multicolumn{1}{c}{Data Availability} & \multicolumn{1}{c}{Code Availability} & \multicolumn{1}{c}{Model}\\ \hline 
\endhead

\hline \multicolumn{5}{r}{{Continued on next page}} \\ \hline
\endfoot

\hline \hline
\endlastfoot

\citep{li2020predicting} & ClinVar \citep{landrum2016clinvar} & \url{https://www.ncbi.nlm.nih.gov/clinvar/} & \url{https://github.com/xzenglab/BertVS} & Y\\
\citep{rives2021biological} & \makecell{\citep{gray2018quantitative},\\\citep{riesselman2018deep}} & & \url{https://github.com/facebookresearch/esm} & Y\\
\citep{meierlanguage} & \citep{suzek2007uniref} (UniRef90) & & \url{https://github.com/facebookresearch/esm} & Y \\
\citep{amadeus2021design} & \citep{yusuf2021genetic} & & & N\\
\citep{avsec2021effective} & \makecell{MPRA: \citep{shigaki2019integration},\\eQTL: Original} & MPRA: \url{http://www.genomeinterpretation.org/cagi5-regulation-saturation.html}, eQTL: \url{https://tinyurl.com/29nafrsw} & \url{https://github.com/google-deepmind/deepmind-research/tree/master/enformer} & Y\\
\citep{ji2021dnabert} & \citep{sherry2001dbsnp} & & \url{https://github.com/jerryji1993/DNABERT} & Y\\
\citep{yamaguchi2021evotuning} & N/A & N/A & \url{https://github.com/dlnp2/evotuning_protocols_for_transformers} & N\\
\citep{liu2022protein}* & ClinVar \citep{landrum2016clinvar} & \url{https://www.ncbi.nlm.nih.gov/clinvar/} & N/A & N\\
\citep{marquet2021embeddings} & \makecell{\citep{hecht2015better},\\\citep{riesselman2018deep},\\\citep{reeb2020variant}} & \url{https://zenodo.org/records/5238537} & \url{https://github.com/Rostlab/VESPA} & Y\\
\citep{yang2022integrating} & \citep{rentzsch2019cadd} & \url{https://figshare.com/articles/dataset/LOGO_dbSNP_score_chr/19149827/2} & \url{https://github.com/melobio/LOGO} & Y\\
\citep{olenyi2022lambdapp:} & \makecell{\citep{apweiler2004uniprot}\\- accession: Q9NZC2} & \url{https://www.uniprot.org/} & \url{https://github.com/Rostlab/LambdaPP/tree/main} & N\\
\citep{zhou2022unsupervised}* & ClinVar: \citep{landrum2016clinvar} & Public Domain & N/A & N\\
\citep{manfredi2022e-snpsgo:} & \makecell{ClinVar: \citep{landrum2016clinvar},\\UniProt: \citep{apweiler2004uniprot}} & \url{https://esnpsandgo.biocomp.unibo.it/datasets/} & N/A & N \\
\citep{dampier2022hiv} & Original & \url{https://huggingface.co/damlab} & Code: \url{https://github.com/DamLabResources/hiv-transformers} & Y\\
\citep{jiang2023deciphering} & HGMD: \citep{stenson2020human} & HGMD professional version required. & \url{https://github.com/WGLab/MutFormer} & Y\\
\citep{sun2023structure-informed}* & \makecell{\citep{riesselman2018deep},\\MSA: \citep{suzek2007uniref} (UniRef100)} & & \url{https://github.com/Stephen2526/Structure-informed_PLM} & Y\\
\citep{brandes2023genome-wide} & \makecell{ClinVar: \citep{landrum2016clinvar},\\HGMD: \citep{stenson2020human},\\\citep{karczewski2020mutational}} & \url{https://github.com/ntranoslab/esm-variants} & \url{https://github.com/ntranoslab/esm-variants} & Y\\
\citep{fan2022shine:} & \citep{chang2018accelerating, kaplanis2020evidence} &  \url{https://github.com/xf-omics/SHINE} & \url{https://github.com/xf-omics/SHINE} & Y\\
\citep{benegas2023gpn-msa}* & \makecell{ClinVar: \citep{landrum2016clinvar},\\\citep{karczewski2020mutational}} & \url{https://huggingface.co/collections/songlab/gpn-msa-65319280c93c85e11c803887} & \url{https://github.com/songlab-cal/gpn} & Y\\
\citep{derbel2023accurate} & N/A & N/A & \url{https://github.com/qgenlab/Rep2Mut} & N\\
\citep{hidayat2023utilizing} & \makecell{ClinVar: \citep{landrum2016clinvar},\\UniProt: \citep{apweiler2004uniprot}} & Public Domain & N/A & N\\
\citep{james2023deep} & ClinVar: \citep{landrum2016clinvar} & Public Domain & \url{https://github.com/facebookresearch/esm}, \url{https://github.com/OATML/EVE} & Y\\
\citep{zhou2024dnabert-2:}* & \citep{khare2021gisaid} & \url{https://github.com/MAGICS-LAB/DNABERT_2} & \url{https://github.com/MAGICS-LAB/DNABERT_2} & Y\\
\citep{cheng2023accurate} & \makecell{\citep{landrum2016clinvar},\\\citep{sundaram2018predicting}, \\\citep{notin2022tranception}, Original curated benchmark} & \url{https://github.com/OATML-Markslab/Tranception}, \url{https://github.com/google-deepmind/alphafold/tree/main/afdb} & \url{https://github.com/google-deepmind/alphamissense} & N\\
\citep{danzi2023deep} & \makecell{ClinVar: \citep{landrum2016clinvar},\\\citep{karczewski2020mutational}} & Public Domain & \url{https://github.com/ZuchnerLab/Maverick} & Y\\
\citep{truongpoet:} & \citep{notin2022tranception} & \url{https://github.com/OATML-Markslab/Tranception} & \url{https://github.com/OpenProteinAI/PoET} & Y\\
\citep{qu2023ensemble} & \citep{notin2022tranception} & \url{https://github.com/OATML-Markslab/Tranception} & N/A & N\\
\citep{dalla2024nucleotide} & \makecell{ClinVar: \citep{landrum2016clinvar},\\HGMD: \citep{stenson2020human},\\\citep{national2021grasp}} & See 'Data Availability' section in paper & \url{https://github.com/instadeepai/nucleotide-transformer} & Y\\
\citep{blaabjerg2024ssemb} & \citep{notin2022tranception} & \url{https://zenodo.org/records/12798019} & \url{https://github.com/KULL-Centre/_2023_Blaabjerg_SSEmb} & Y\\
\citep{lin2024enhancing} & \makecell{ClinVar: \citep{landrum2016clinvar},\\\citep{karczewski2020mutational}} & Public Domain & \url{https://github.com/wlin16/VariPred} & Y\\
\citep{wild2024dna}* & \makecell{ClinVar: \citep{landrum2016clinvar},\\\citep{lambert2021polygenic}} & N/A & N/A & N\\
\citep{luo2024interpretable} & \makecell{\citep{chuai2018deepcrispr},\\\citep{lin2020crispr}} & N/A & \url{https://github.com/BrokenStringx/CRISPR-BERT} & Y\\
\citep{gao2023epigept:}* & \makecell{ClinVar: \citep{landrum2016clinvar},\\\citep{wang2021leveraging}} & See links in paper & \url{https://github.com/ZjGaothu/EpiGePT} & Y\\
\citep{zhan2024dyna:}* & N/A & N/A & \url{https://github.com/zhanglab-aim/DYNA} & Y\\
\citep{lafita2024machine}* & ClinVar: \citep{landrum2016clinvar} & Public Domain & N/A & N\\
\citep{marquet2024vespag:} & \citep{notin2022tranception} & \url{https://github.com/OATML-Markslab/Tranception} & \url{https://github.com/JSchlensok/VespaG} & Y\\
\citep{yan2024transefvp:} & \citep{manfredi2022e-snpsgo:} & \url{https://github.com/yzh9607/TransEFVP/tree/master} & \url{https://github.com/yzh9607/TransEFVP/tree/master} & N \\
\citep{Shulgina2024.04.05.588317}* & Original & \url{https://tinyurl.com/5abszup9} &  \url{https://github.com/Doudna-lab/GARNET_DL} & Y \\
\citep{li2024mvformer} & Original & \url{https://github.com/genemine/MVFormer} & \url{https://github.com/genemine/MVFormer} & N\\
\citep{zhong2024premode}* & Original & \url{https://huggingface.co/gzhong/PreMode} & \url{https://github.com/ShenLab/PreMode} & Y\\
\citep{linder2025predicting} & N/A & \url{gs://borzoi-paper/data/} & \url{https://github.com/calico/borzoi} & Y\\
\citep{joshi2025augmented} & UniProt: \citep{apweiler2004uniprot} & Public Domain & Available upon request & N\\
\citep{glaser2025esm}* & N/A & N/A & \url{https://github.com/moritzgls/ESM-Effect} & N\\

\end{longtable}

\begin{table*}[h!]
\caption{Summary of post-Transformer language models for variant effect prediction. * = preprint.\label{supp:post_transformer_models}}
\tabcolsep=5pt%
\begin{tabular*}{\linewidth}{@{\extracolsep{\fill}}llp{0.25\textwidth}p{0.4\textwidth}l@{\extracolsep{\fill}}}
\toprule%
Paper & Data Source & Data Availability & Code Availability & Model\\
\midrule
\citep{schiff2024caduceus:}* & eQTL: \citep{avsec2021effective} & \url{https://tinyurl.com/29nafrsw} & \url{https://github.com/kuleshov-group/caduceus} & Y\\
\citep{nguyen2024evo} & \makecell{\citep{notin2023proteingym},\\\citep{kobori2015high}} & N/A & Code: \url{https://github.com/evo-design/evo} & Y\\
\bottomrule
\end{tabular*}
\end{table*}

\begin{table*}[ht!]
\caption{Links to the benchmarks summarised in Table \ref{tab:benchmarks}.\label{supp:benchmarks}}
\tabcolsep=5pt%
\begin{tabular*}{\linewidth}{@{\extracolsep{\fill}}p{0.5\textwidth}p{0.5\textwidth}@{\extracolsep{\fill}}}
\toprule%
Benchmark & Link \\
\midrule
Benchmarking of variant effect predictors using deep mutational scanning \citep{livesey2020using} & \makecell{\url{https://doi.org/10.6084/m9.figshare.12369359.v1},\\
\url{https://doi.org/10.6084/m9.figshare.12369452.v1}}\\
BEND \citep{marin2023bend} & \url{https://github.com/frederikkemarin/BEND}\\
Updated benchmarking of variant effect predictors using deep mutational scanning \citep{livesey2023updated} & \url{https://figshare.com/articles/dataset/Compiled_DMS_and_VEP_predictions/21581823/1}\\
Genome Understanding Evaluation \citep{zhou2024dnabert-2:} & \url{https://github.com/Zhihan1996/DNABERT_2}\\
Genomic Long-Range Benchmark \citep{kao2024advancing} & \url{https://huggingface.co/datasets/InstaDeepAI/genomics-long-range-benchmark}\\
\bottomrule
\end{tabular*}
\end{table*}

\end{landscape}

\end{document}